\documentclass[12pt]{cmuthesis}

\usepackage[numbers,sort]{natbib}
\usepackage{ifthen}
\usepackage[pagestyles,extramarks]{titlesec}

\usepackage{float}
\usepackage{times}
\usepackage{fullpage}
\usepackage[pdftex]{graphicx}
\usepackage{amssymb,amsfonts,amsmath,amscd}
\usepackage{gensymb}

\usepackage{amsthm} % for new theorem
\usepackage{thmtools} 
\usepackage{mathtools}
\usepackage{epstopdf}
\usepackage{lipsum}  
\usepackage{caption}%needed for subcaption
\usepackage{subcaption} 
\usepackage{units}
\usepackage{booktabs} %for table rulers
\usepackage{tabulary}
\usepackage{pdflscape}
\usepackage{afterpage}
\usepackage[pass]{geometry}
\usepackage{multirow}

\usepackage{pifont}% http://ctan.org/pkg/pifont

\usepackage{parskip}
\usepackage[usenames,dvipsnames]{color} %Options expand the named colours
\usepackage{soul}
\usepackage{nicefrac}
\usepackage{enumitem}
\usepackage{xspace}
\usepackage{pdflscape}
\usepackage{afterpage}

\usepackage{layouts}
\usepackage{rotating}
\usepackage{csquotes}

\usepackage[nolist]{acronym} %updated version of acronym.sty is in ./ folder.
\usepackage[toc]{appendix}

\usepackage[font=footnotesize]{caption}%needed for subcaption
\usepackage[font+=small]{subcaption} 

\usepackage{algorithm}
\usepackage{algpseudocode}% http://ctan.org/pkg/algorithmicx

\algtext*{EndWhile}% Remove "end while" text
\algtext*{EndIf}% Remove "end if" text
\algtext*{EndFor}% Remove "end if" text

\usepackage[pdftex, backref,pageanchor=true,plainpages=false, pdfpagelabels, bookmarks,bookmarksnumbered,
]{hyperref}

\settitlemarks*{section,subsection}

\newpagestyle{mystyle}{
\headrule\footrule
\sethead{}{\thesection.\ \sectiontitle\ifthesubsection{\ --\ \itshape\firstextramarks{subsection}\subsectiontitle}{}}{}
\setfoot{}{\thepage}{}
}
\makeatletter
\def\th@plain{%
  \thm@notefont{}% same as heading font
  \itshape % body font
}
\def\th@definition{%
  \thm@notefont{}% same as heading font
  \normalfont % body font
}
\makeatother

\makeatletter
\renewcommand*\env@matrix[1][\arraystretch]{%
  \edef\arraystretch{#1}%
  \hskip -\arraycolsep
  \let\@ifnextchar\new@ifnextchar
  \array{*\c@MaxMatrixCols c}}
\makeatother

\makeatletter
\newcommand*\bigcdot{\mathpalette\bigcdot@{.5}}
\newcommand*\bigcdot@[2]{\mathbin{\vcenter{\hbox{\scalebox{#2}{$\m@th#1\bullet$}}}}}
\makeatother

\graphicspath{{../images/}}

\begin {document} 
\frontmatter

%initialize page style, so contents come out right (see bot) -mjz
\pagestyle{empty}

\title{  {\it \huge }\\
{\bf Toward Increased Airspace Safety: Quadrotor Guidance for Targeting Aerial Objects}}
\author{Anish Bhattacharya}
\date{August 2020}
\Year{2020}

\trnumber{CMU-RI-TR-20-39}

\committee{
Sebastian Scherer (Chair) \\
Oliver Kroemer\\
Azarakhsh Keipour
}

\support{}
\disclaimer{}

% copyright notice generated automatically from Year and author.
% permission added if \permission{} given.
%\keywords{}

\maketitle

\begin{dedication}
To my parents and my girlfriend for their constant support.
\end{dedication}

\pagestyle{plain} % for toc, was empty

\begin{abstract}

% As the market for commercially available unmanned aerial vehicles (UAVs) booms, there is an increasing number of small, teleoperated or autonomous aircraft found in protected or sensitive airspace. Operations to remove offending aircraft can be costly and extremely disruptive. The high maneuverability of quadrotor UAVs offers a potential solution to targeting and removing aircraft with unknown flight characteristics. In this work, multiple methods are described and evaluated for guidance of a quadrotor in impacting moving targets with unknown trajectories. In addition, the system developed for the CMU Team Tartan entry in the MBZIRC 2020 Challenge 1 competition, focused on targeting both stationary and moving objects, is covered. Real-world results are included from the competition.

As the market for commercially available unmanned aerial vehicles (UAVs) booms, there is an increasing number of small, teleoperated or autonomous aircraft found in protected or sensitive airspace. Existing solutions for removal of these aircraft are either military-grade and too disruptive for domestic use, or compose of cumbersomely teleoperated counter-UAV vehicles that have proven ineffective in high-profile domestic cases. In this work, we examine the use of a quadrotor for autonomously targeting semi-stationary and moving aerial objects with little or no prior knowledge of the target's flight characteristics. Guidance and control commands are generated with information just from an onboard monocular camera. We draw inspiration from literature in missile guidance, and demonstrate an optimal guidance method implemented on a quadrotor but not usable by missiles. Results are presented for first-pass hit success and pursuit duration with various methods. Finally, we cover the CMU Team Tartan entry in the MBZIRC 2020 Challenge 1 competition, demonstrating the effectiveness of simple line-of-sight guidance methods in a structured competition setting.

\end{abstract}

\begin{acknowledgments}

I would like to thank my advisor, Prof. Sebastian Scherer, for his support and guidance throughout the work involved in this thesis. I would also like to thank my committee members, Prof. Oliver Kroemer and Azarakhsh Keipour for their consideration and feedback. The advice, mentorship, and friendship provided by the members of the AirLab and greater RI was a key element to my experience at CMU; specifically, I would like to recognize Rogerio Bonatti, Azarakhsh Keipour, and Vai Viswanathan for their help and guidance. John Keller was extremely helpful for setting up DJI SDK integration as well as advising with random software challenges. Finally, I would like to thank the members of CMU Team Tartans: Akshit Gandhi, Noah LaFerriere, Lukas Merkle, Andrew Saba, Rohan Tiwari, Stanley Winata, and Karun Warrior. Jay Maier, Lorenz Stangier, and Kevin Zhang also contributed to this effort.

\end{acknowledgments}

\tableofcontents

\listoffigures
\listoftables

\mainmatter

%% Double space document for easy review:
%\renewcommand{\baselinestretch}{1.66}\normalsize

% The other requirements Catherine has:
%
%  - avoid large margins.  She wants the thesis to use fewer pages, 
%    especially if it requires colour printing.
%
%  - The thesis should be formatted for double-sided printing.  This
%    means that all chapters, acknowledgements, table of contents, etc.
%    should start on odd numbered (right facing) pages.
%
%  - You need to use the department standard tech report title page.  I
%    have tried to ensure that the title page here conforms to this
%    standard.
%
%  - Use a nice serif font, such as Times Roman.  Sans serif looks bad.
%
% Other than that, just make it look good...

\chapter{Introduction}
\label{chap:introduction}

\section{Motivation}

\begin{figure}[h!]
\centering
\subcaptionbox{\label{fig:drone-net}Iterations of the South Korean police counter-UAV aerial system equipped with nets \cite{mailonline_2018}.}{\includegraphics[height=5.5cm]{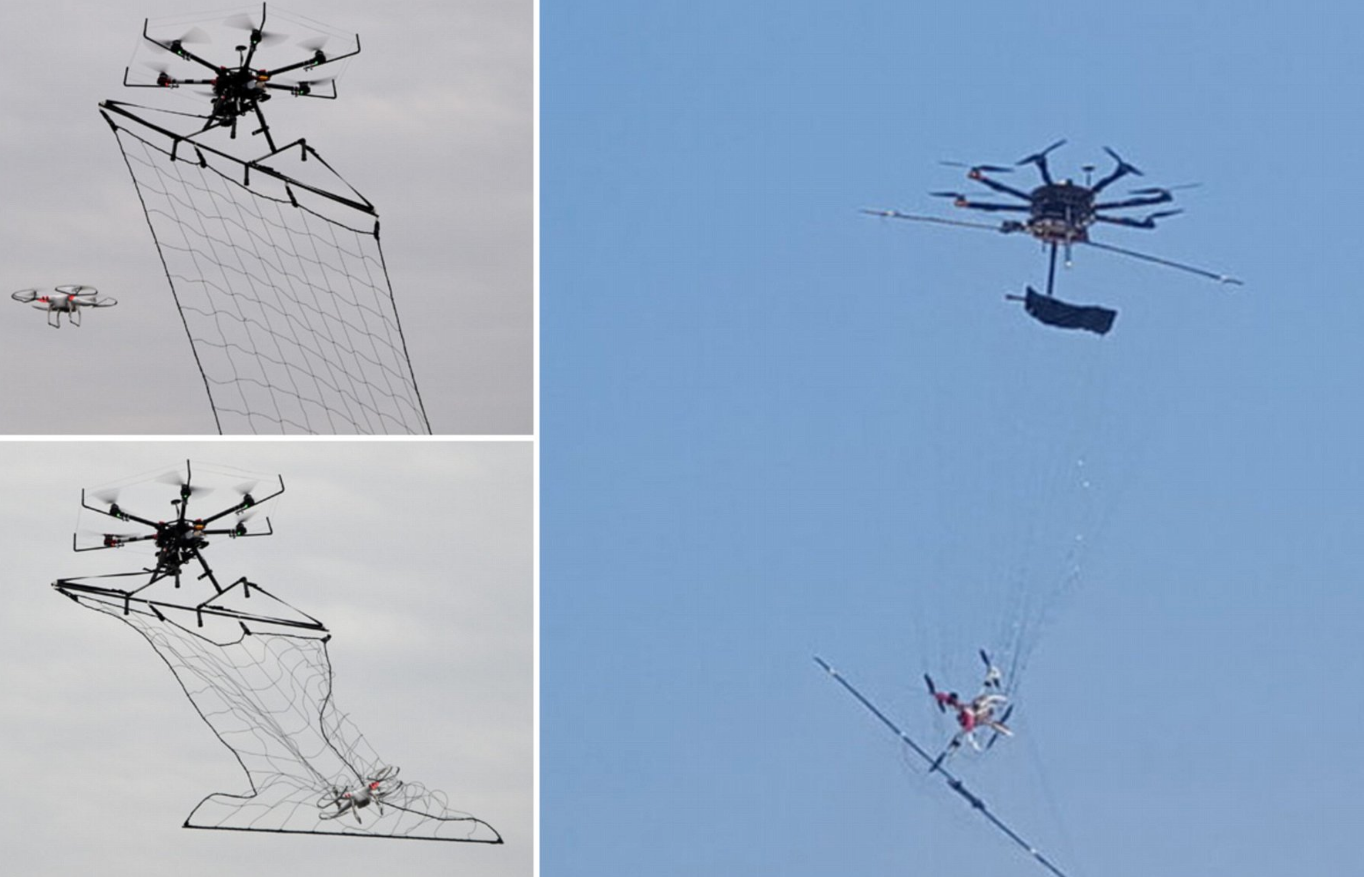}}\hfill
\subcaptionbox{\label{fig:anduril}Anduril counter-UAV (lower) utilizing a ``battering ram" approach to disable a target \cite{brandom_2019}.}{\includegraphics[height=5.5cm]{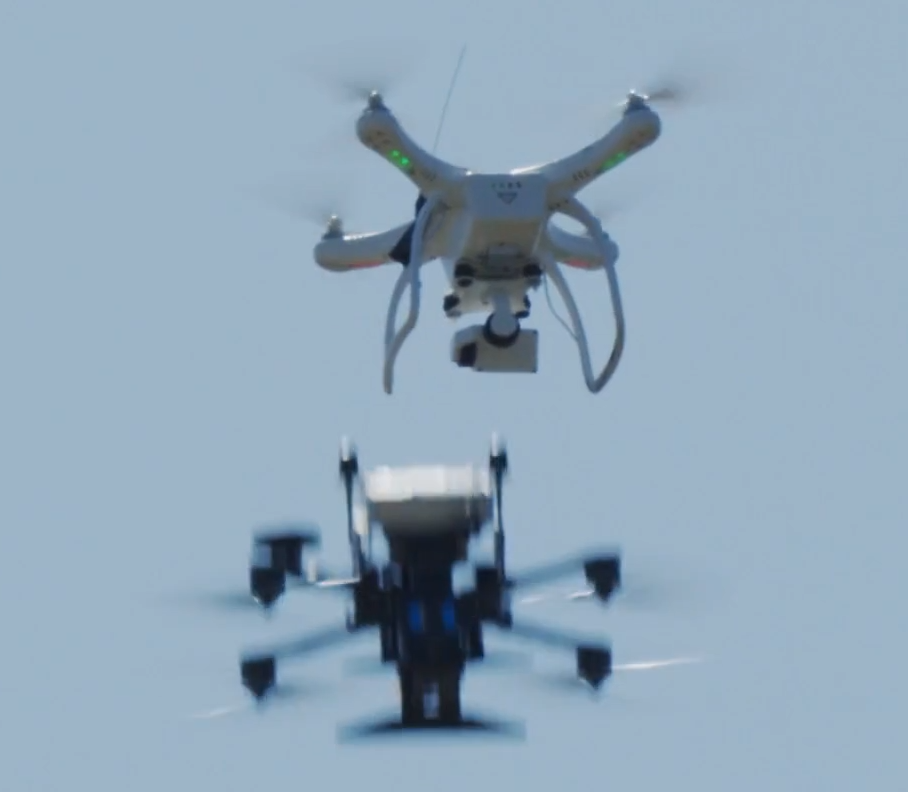}}\\
\caption{Current possible domestic-use counter-UAV systems in development.}
\label{fig:current-cUAVs}
\end{figure}

Micro Aerial Vehicles (MAVs), also referred to as drones, Unmanned Aerial Vehicles (UAVs) and small Unmanned Aerial Systems (sUAS), have seen a huge growth in various market sectors across the globe. Business Insider projects the sale of drones to surpass \$12 billion in 2021, of which consumer drone shipments will comprise 29 million units \cite{intelligence_2020}. Enterprise and governmental sectors generally have strict regulations under which MAVs are operated; however, not only is the consumer sector's operation of these aircraft weakly regulated but there is strong community pushback against any such legislation. Stronger oversight of private drone use is further motivated by numerous incidents involving small, typically teleoperated drones in public spaces. In December 2018, dozens of drone sighting reports over Gatwick Airport, near London, affected 1,000 flights and required the help of both the police and military, neither of whom were able to capture the drone over a 24-hour period \cite{mckenzie_mezzofiore_2018}. Worries over the potential of UAVs above crowds at the 2018 Winter Olympics prompted South Korean authorities to train to disable drones, including developing a teleoperated platform which disables others with a dropped net (Figure \ref{fig:drone-net}) \cite{mailonline_2018}. Beyond these documented examples, there are numerous videos online of recreational drone users losing control of their aircraft due to weather conditions, loss of GPS positioning, or low battery behavior.

% \begin{figure}[h!]
%     \centering
%     \begin{minipage}{.55\textwidth}
%     \centering
%     \includegraphics[height=5cm]{images/drone-net.png}
%     \caption{Iterations of the South Korean police counter-UAV aerial system equipped with nets \cite{mailonline_2018}}
%     \label{fig:drone-net}
%     \end{minipage}%
%     \begin{minipage}{.45\textwidth}
%     \centering
%     \includegraphics[height=5cm]{images/anduril.png}
%     \caption{Anduril counter-UAV (lower) utilizing a ``battering ram" approach to disable a target \cite{brandom_2019}}
%     \label{fig:anduril}
%     \end{minipage}
% \end{figure}

While these issues could be mitigated by enforcing strict regulations and oversight on the consumer drone market, this may also drastically curb the independence of hobbyists and researchers. A potential alternative may be to capture or disable rogue UAVs in a non-disruptive way. Current anti-UAV technology exists primarily in the military sector, in the form of jammers (used to disrupt the teleoperation signal from a nearby radio controller) or Stinger missiles (meant to disable a UAV by impact). Neither of these options are suitable for domestic use, where both noise and debris are of issue. Therefore, we need a solution that minimizes destruction while being agile enough to capture a wide variety of MAVs in local airspaces. Quadrotors benefit from high, 4-degree of freedom (DOF) maneuverability and can accelerate to high speeds quicker than some single-rotor or fixed-wing counterparts. This implies a higher capability to stay on an impact trajectory towards a target with an unknown flight plan or characteristics. Furthermore, recent research in aerial robotics has shown that a suite of obstacle avoidance, detection, planning, and control can run fully onboard on an autonomous quadrotor platform. Common shortfalls of quadrotors include low battery life, but for this mission type, flights are short but with high accelerations (and therefore, higher energy throughput).

\section{Challenges and Approach}

The challenges of autonomously impacting an unknown small aerial vehicle with a UAV are numerous, involving fast flight through potentially cluttered environments, as well as the development of a mechanically sound method of capturing the target without damage to either agent. However, the primary challenge addressed in this thesis surrounds guidance and control of a quadrotor UAV towards a target.

% \begin{enumerate}
%     \item Control and guidance of a UAV towards a target with unknown trajectory or flight characteristics.
%     \item Accurate detection and tracking of a small target against a noisy background.
% \end{enumerate}

% TODO insert picture of small drone against background

In this thesis, we describe two projects to address this challenge. In the first study, multiple control and guidance methods derived and inspired from different fields of literature are reviewed and modified for use on a simulated UAV-target scenario. Here, the perception task is simplified and environmental factors are eliminated to focus on the evaluation of several guidance methods. The second study evaluates LOS guidance in a robotics competition setting, specifically comprising of the CMU Team Tartan entry in the Mohamed Bin Zayed International Robotics Challenge 2020 Challenge 1. This effort includes work on (a) planning an adjustable and robust path around a fixed arena based on measured GPS coordinates, (b) control towards semi-stationary targets placed throughout the arena, and (c) detection of a small yellow ball moving at 8m/s against a cluttered background.

A further challenge in this work is the localization of the target in the world relative to the UAV. Depending on the size and shape of target (e.g. fixed wing, multirotor, helicopter, blimp) as well as its distance, it cannot be assumed that a 3D sensor, such as LIDAR or stereo vision, can be used to accurately localize the target in space. For example, because of their sparse and sometimes mesh-structured frames, multirotors in particular can be notoriously difficult to localize with cheap and lightweight scanning LIDARs or stereo cameras at long range. Therefore, the focus in this thesis is to use monocular vision and adapt guidance methods to use only approximate depth estimates when necessary.

\section{Contribution and Outline}

The main contributions of this thesis are as follows.

\begin{enumerate}
    \item An evaluation of various guidance and control methods and how they might be adapted for use on a quadrotor in a simulated environment.
    \item A software system using LOS-guidance for finding and targeting semi-stationary and moving targets within a fixed arena.
\end{enumerate}

Chapter \ref{chap:related_work} presents a short summary of related work in various fields, including classical visual servoing, missile guidance, and trajectory generation and tracking. The following two chapters, \ref{chap:moving_objects} and \ref{chap:mbzirc}, expand on the work done specifically towards the two contributions listed above, respectively. Chapter \ref{chap:conclusion} describes conclusions drawn from this work, shortcomings of the approach, as well as suggested future directions.

\chapter{Related Work}
\label{chap:related_work}

\section{Classical Visual Servoing}
\label{sec:rw-vis_servo}

Visual servoing spans a wide range of research focusing on controlling robot links relative to input from visual sensors. The most common application of visual servoing is in pick-and-place operations done with robotic arms fitted with cameras. These robots generally either have a eye-in-hand (closed-loop control) or eye-to-hand (open-loop control) setup \cite{Chaumette07a}. Two of the most common approaches in this field are image-based visual servoing (IBVS) and pose-based visual servoing (PBVS), with the difference between the two involving the estimation of the target's relative pose to the robot \cite{hutchinson1996tutorial}.

IBVS, as described in Hutchinson, et al. (1996), is only useful within a small region of the task space unless the image Jacobian is computed online with knowledge of the distance-to-target, which is further complicated by a monocular vision-based system. Unless target velocity is constant, errors or lag in the image plane with a moving target introduces errors in servoing with either method, which would in turn have to be tuned out with a more complex control system. Chaumette and Santos (1993) \cite{chaumette_santos_1993} tracked a moving target with IBVS but assumed a constant acceleration. When the target maneuvered abruptly, the Kalman filter-based target motion predictor took some cycles of feedback to recalibrate to the new motion. In \cite{corke2000real}, the major pitfall of PBVS is pointed out as the need for 3D information of the target, specifically the depth which may not be readily available.

\section{Missile Guidance}
\label{sec:rw-missile}

\begin{figure}[h!]
    \begin{minipage}{.45\textwidth}
    \centering
    \includegraphics[width=0.9\linewidth]{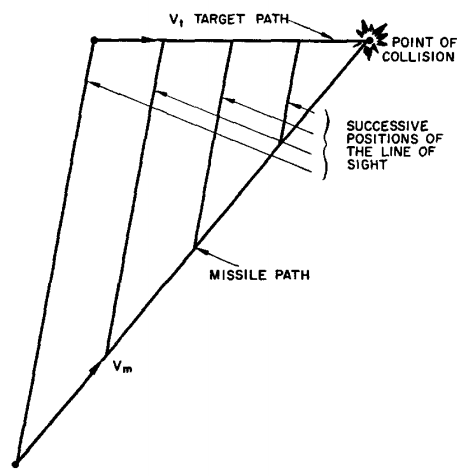}
    \caption{Constant-bearing collision course diagram with instantaneous lateral acceleration applied to missile \cite{adler1956missile}.}
    \label{fig:constant-bearing}
    \end{minipage}%
    \centering
    \begin{minipage}{.55\textwidth}
    \centering
    \includegraphics[width=0.7\linewidth]{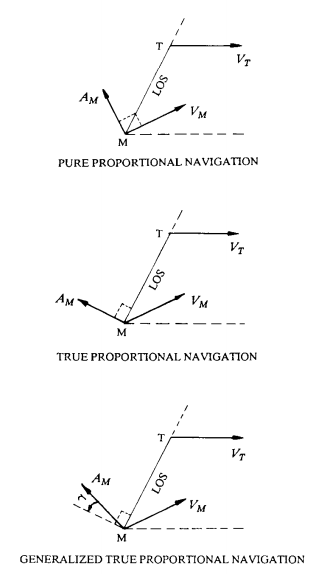}
    \caption{Comparison of pure and true proportional navigation. $A_M$ and $V_M$ refer to the missile's desired acceleration and current velocity, respectively \cite{shukla1990proportional}.}
    \label{fig:pn-comparisons}
    \end{minipage}
\end{figure}

Homing air missiles are singularly focused on ensuring impact with an aerial target. Since at least 1956, proportional navigation in some form has been a standard in missile guidance methods \cite{adler1956missile}. Adler (1956) describes a couple of such methods, including constant-bearing navigation and proportional navigation. It is noted that constant-bearing navigation, which applies acceleration to maintain a constant bearing-to-target, requires instantaneous corrections to deviations in the line-of-sight (LOS) direction. This renders it incompatible with the dynamics of missiles, which cannot directly satisfy lateral acceleration commands (similar to fixed-wing aircraft); therefore, Adler proposes using 3D proportional navigation (PN) which applies a turning rate proportional to the LOS direction change. In later texts, the term proportional navigation is used interchangeably between these two schemes, and also extended to other similar methods. In this thesis, PN will be used as a general term to refer to any control law using the LOS rotation rate to generate an acceleration command. As noted in \cite{palumbo2010basic}, PN, when assuming no autopilot lag, is an optimal control law that assumes very little about the acceleration characteristics of the target. However, variations on classical PN have also been developed that adapt to different flight profiles, including constant-acceleration and constant-jerk \cite{palumbo2010modern}. PN is typically split into two categories, the ``true" variant and the ``pure" variant \cite{shukla1990proportional}. Though the naming is largely arbitrary, the primary difference lies in the reference frame in which the lateral acceleration is applied to the pursuing missile. True PN applies this acceleration orthogonal to the current missile velocity; Pure PN applies the acceleration orthogonal to the current LOS towards the target. Generalized True PN (as seen in Figure \ref{fig:pn-comparisons} from \cite{shukla1990proportional}) is not covered in this thesis.

% TODO include existing work on applying such methods to quadrotors

\section{Trajectory Generation and Tracking}
\label{sec:rw-trajectory}

\begin{figure}[h!]
    \centering
    \includegraphics[width=0.7\linewidth]{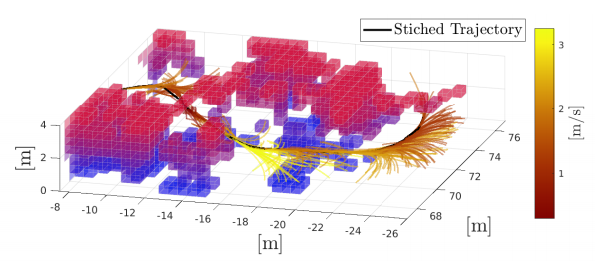}
    \caption{Example of concatenated trajectories in a cluttered environment. Colored lines represent individual motion primitives; the quadrotor tracks the initial part of every generated trajectory, forming a complete, stitched trajectory represented by the black line. \cite{ryll2019efficient}}
    \label{fig:stitched-trajectories}
\end{figure}

Trajectories provide robots with smooth or kinodynamically feasible paths to follow through its state space. This is opposed to sending raw differential commands, which may exceed the robot's limitations and lead to controller instability or even to physical breakdown of the robot's actuators. As such, there has been extensive work in the generation and following of trajectories for use with various types of robots and applications, primarily with robot arms for grasping and self-driving vehicles \cite{stolle2006policies}\cite{frazzoli2000robust}\cite{dey2012efficient}\cite{berenson2007grasp}. This has been extended to MAVs to ensure smooth and efficient flight. Richter, et al. (2013) \cite{richter2016polynomial} showed that polynomial trajectories eliminated the need for an extensive sample-based search over the state space of the vehicle. This approach, while not providing asymptotic convergence to the global optimal path, ensured smooth and fast flight of the quadrotor. In \cite{ryll2019efficient}, it was shown that with continuous minimum-jerk trajectories and fast re-planning, they achieved higher trajectory smoothness compared to other, more reactive planners. Figure \ref{fig:stitched-trajectories} shows the smooth trajectory generated by tracking motion primitive-generated paths. Gao, et al. (2018) \cite{gao2018online} first finds a time-indexed minimal arrival path that may not be feasible for quadrotor flight, and then forms a surrounding free-space flight corridor in which they generate a feasible trajectory. They use a Bernstein polynomial basis and represent the trajectory as a piecewise B\'{e}zier curve.

To follow trajectories, controllers take in a desired trajectory typically composed of position waypoints each with an associated velocity, and issue actuator commands to the robot. In the MAV case, an autopilot software may accept attitude or attitude-rate commands which come from such a controller. Hoffman, et al. (2008) \cite{hoffmann2008quadrotor} demonstrated a controller that took as input a non-feasible trajectory and outputted feasible attitude commands for a quadrotor that accurately followed the original path. This was demonstrated outdoors with 50cm accuracy. A similar approach was taken (but extended to 3D) for the path following controller implemented in \cite{mellinger2012trajectory}. Here, the desired position and desired velocity from the closest point on the trajectory are used to find the position and velocity errors of the robot. These are used to calculate the desired robot acceleration with PD feedback and a feedforward term.

 \chapter{Targeting Moving Objects with a Quadrotor}
\label{chap:moving_objects}

% Purpose: evaluate various methods in gazebo simulation

% gazebo simulation: environment, including simplifications (no obstacles, wind, etc). target trajectory track types shown in RVIZ.

% system details, including PX4 and controllers, perception, etc

As described in Section \ref{sec:rw-vis_servo}, servoing towards moving targets is challenging with classical methods. As such, LOS-based guidance principals (Section \ref{sec:rw-missile}) and trajectory following methods (\ref{sec:rw-trajectory}) may produce better results when target acceleration is nonzero. In addition, the quadrotor platform's control limits might be avoided with smooth trajectory-based methods. This chapter focuses on the development and evaluation of various such guidance methods to achieve impact with a generalized form of an aerial, mobile target. No information is known about the target other than its color, which is used for segmentation in an RGB image to simplify the detection and tracking problem.

\section{Line-of-Sight Guidance}
\label{sec:los-guidance}

\subsection{Derivation of True Proportional Navigation Guidance Law}
\label{subsec:tpn-derivation}

Line-of-sight (LOS) guidance methods are used to apply acceleration commands to the pursuer that minimize change in the LOS vector towards the target. In this section, the basic LOS geometry is introduced and used to derive proportional navigation (PN) guidance. Following subsections show how this is used, with target detections, to calculate quantities used for the applied PN guidance.

\begin{figure}[h!]
    \centering
    \includegraphics[width=0.7\textwidth]{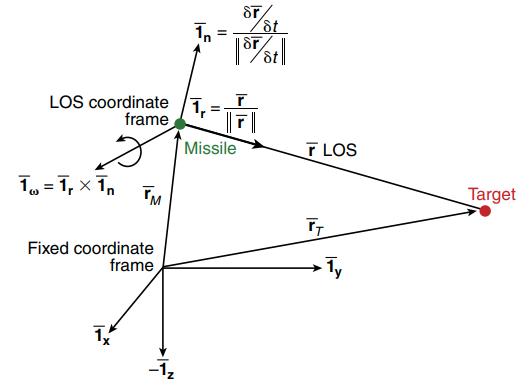}
    \caption{LOS coordinate system \cite{palumbo2010basic}.}
    \label{fig:los-coords}
\end{figure}

As seen in Figure \ref{fig:los-coords}, the fixed world coordinate frame is specified by the unit vectors $\bar{\mathbf{1}}_{\mathbf{x}}$, $\bar{\mathbf{1}}_{\mathbf{y}}$, $\bar{\mathbf{1}}_{\mathbf{z}}$. The LOS coordinate frame, attached to the moving missile, is specified by the unit vectors $\bar{\mathbf{1}}_{\mathbf{r}}$, $\bar{\mathbf{1}}_{\mathbf{n}}$, $\bar{\mathbf{1}}_{\mathbf{\omega}}$; $\bar{\mathbf{1}}_{\mathbf{r}}$ points along the LOS $\bar{\mathbf{r}}$; $\bar{\mathbf{1}}_{\mathbf{n}}$ is the change in direction (i.e. a rotation) of the LOS vector; $\bar{\mathbf{1}}_{\mathbf{\omega}}$ is the cross product of the former two, in that order (forming a right-handed coordinate frame).

In general, the angular velocity of the LOS coordinate frame is given by:

\begin{equation}
    \bar{\dot{\phi}}
    = \dot{\Phi}_{\mathbf{r}} \bar{\mathbf{1}}_{\mathbf{r}}
    + \dot{\Phi}_{\mathbf{n}} \bar{\mathbf{1}}_{\mathbf{n}}
    + \dot{\Phi}_{\mathbf{\omega}} \bar{\mathbf{1}}_{\mathbf{\omega}}
\end{equation}

Where $\dot{\Phi}_{\mathbf{r}}$, $\dot{\Phi}_{\mathbf{n}}$, $\dot{\Phi}_{\mathbf{\omega}}$ are the magnitudes of the components of the angular velocity defined as:

\begin{align}
    \dot{\Phi}_{\mathbf{r}} &= \bar{\dot{\phi}} \bigcdot \bar{\mathbf{1}}_{\mathbf{r}}\\
    \dot{\Phi}_{\mathbf{n}} &= \bar{\dot{\phi}} \bigcdot \bar{\mathbf{1}}_{\mathbf{n}}\\
    \label{eq:phi-omega-dot}
    \dot{\Phi}_{\mathbf{\omega}} &= \bar{\dot{\phi}} \bigcdot \bar{\mathbf{1}}_{\mathbf{\omega}}
\end{align}

As derived in \cite{palumbo2010basic} but not reproduced here, the components of the relative acceleration between the missile and target are:

\begin{align}
    \label{eq:delta-a-first}
    (\bar{\mathbf{a}}_T - \bar{\mathbf{a}}_M) \bigcdot \bar{\mathbf{1}}_{\mathbf{r}}
    &= \Ddot{R} - R \dot{\Phi}_{\mathbf{\omega}}^2\\
    (\bar{\mathbf{a}}_T - \bar{\mathbf{a}}_M) \bigcdot \bar{\mathbf{1}}_{\mathbf{n}}
    &= 2\dot{R}\dot{\Phi}_{\mathbf{\omega}} + R \Ddot{\Phi}_{\mathbf{\omega}}\\
    (\bar{\mathbf{a}}_T - \bar{\mathbf{a}}_M) \bigcdot \bar{\mathbf{1}}_{\mathbf{\omega}}
    &= R \dot{\Phi}_{\mathbf{\omega}} \dot{\Phi}_{\mathbf{r}}
\end{align}

Where $\bar{\mathbf{a}}_T$ and $\bar{\mathbf{a}}_M$ are the target and missile accelerations, respectively. From this result, specifically using the condition in Equation \ref{eq:delta-a-first}, we can list sufficient conditions to satisfy the equation and ensure intercept: (\textit{i}) interceptor is able to achieve an acceleration along the LOS greater than that of the target ($(\bar{\mathbf{a}}_T - \bar{\mathbf{a}}_M) \bigcdot \bar{\mathbf{1}}_{\mathbf{r}} < 0$), (\textit{ii}) the initial rate of change in the range $R$ is negative ($\dot{R} < 0$), which then ensures $\Ddot{R} < 0$ given the first condition, and (\textit{iii}) the rate of change in the LOS is $0$ ($\dot{\Phi}_{\mathbf{\omega}}= 0$). Condition (\textit{i}) depends on the nature of the interceptor and target; condition (\textit{ii}) implies that PN pursuit must be initialized with a positive closing velocity; condition (\textit{iii}) implies that the interceptor must satisfy acceleration commands such that the LOS vector remains constant. Palumbo, et al. (2010) finds the following true PN (TPN) law that ensures system stability:

\begin{equation}
\label{eq:acc-mag}
    \bar{\mathbf{a}}_M \bigcdot \bar{\mathbf{1}}_{\mathbf{n}}
    =
    N V_c \dot{\Phi}_{\mathbf{\omega}}\, ,\, N>2
\end{equation}

Where $N$ is a proportional gain and $V_c$ is the closing velocity. In other words, the interceptor acceleration $\bar{\mathbf{a}}_M$ must have a component, orthogonal to the LOS, proportional to the rotation rate of the LOS as specified in Equation \ref{eq:phi-omega-dot}.

\subsection{True Proportional Navigation}
\label{subsec:tpn-alg}

\begin{figure}[h!]
    \begin{minipage}{.4\textwidth}
    \centering
    \includegraphics[width=0.9\textwidth]{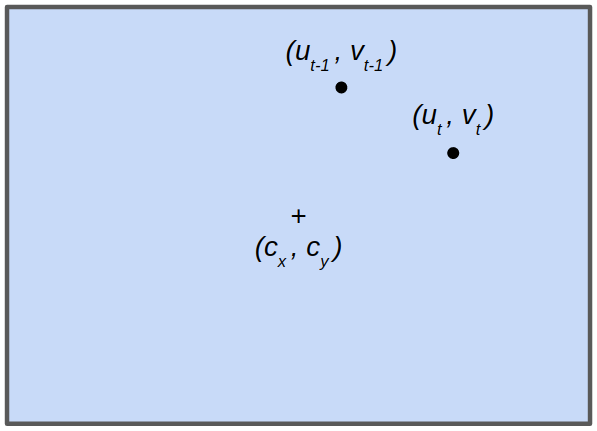}
    \caption{Example image frame with detected object's centroid at times $t-1$ and $t$.}
    \label{fig:image-frame}
    \end{minipage}%
    \centering
    \begin{minipage}{.6\textwidth}
    \centering
    \includegraphics[width=0.9\linewidth]{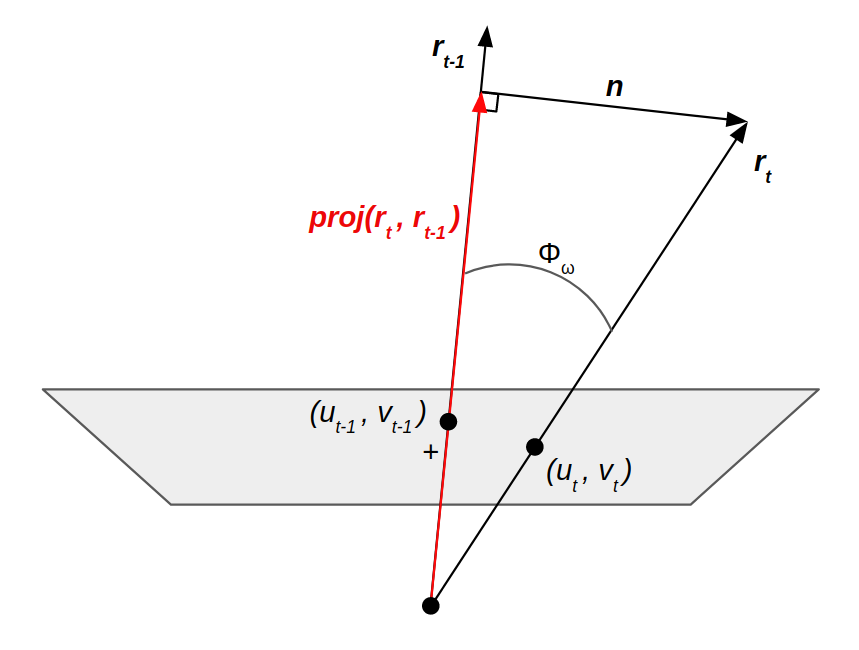}
    \caption{Top-down diagram showing intermediate quantities in calculation of desired acceleration command.}
    \label{fig:pn-calc}
    \end{minipage}
\end{figure}

To generate the desired acceleration vector with magnitude specified by Equation \ref{eq:acc-mag} and direction orthogonal to the LOS, we first calculate both $\dot{\Phi}_{\mathbf{\omega}}$ (directly represented in Equation \ref{eq:acc-mag}) and $\mathbf{1_{\mathbf{n}}}$ (acceleration direction). The target's centroid in image frame coordinates at times $t-1$ and $t$ is represented by $(u_{t-1},v_{t-1})$ and $(u_{t},v_{t})$, respectively, as shown in Figure \ref{fig:image-frame}. The camera principal point, specified in the calibrated camera's intrinsic matrix (typically denoted $\mathbf{K}$), is represented by $(c_x,c_y)$.

The LOS vector $\mathbf{r_t}$ in the camera's frame of reference at time $t$ is given by the following.

\begin{equation}
\label{eq:los}
    \mathbf{r_t} = \begin{bmatrix}[1.75] \dfrac{u_t-c_x}{f_x} \\ \dfrac{v_t-c_y}{f_y} \\ 1 \end{bmatrix}
\end{equation}

In the special case of the first iteration of the algorithm, at $t=0$, the LOS vector is calculated according to Equation \ref{eq:los} then stored for use as $\mathbf{r_{t-1}}$ the upcoming iteration. For the $t=0$ computation cycle the control output is set to $\mathbf{0}$.

The angle spanned by the two vectors $\mathbf{r_{t}}$ and $\mathbf{r_{t-1}}$ is as follows:

\begin{equation}
    \Phi_{\mathbf{\omega}} = \arccos{\dfrac{ <\mathbf{r_{t}},\mathbf{r_{t-1}}> }
    { ||\mathbf{r_{t}}||||\mathbf{r_{t-1}}|| }}
\end{equation}

Therefore, if the difference in time for one cycle is represented by $\Delta t$, then the magnitude of the rotation rate of the LOS vector is:

\begin{equation}
\label{eq:phi-omega-dot}
    \dot{\Phi}_{\mathbf{\omega}} = \frac{\Phi_{\mathbf{\omega}}}{\Delta t}
\end{equation}

The direction of the acceleration (direction of the LOS rotation) $\mathbf{1_{\mathbf{n}}}$ is shown in Figure \ref{fig:pn-calc} and calculated below.

\begin{align}
    \mathbf{n} &= \mathbf{r_{t}} - \texttt{proj}(\mathbf{r_{t}}, \mathbf{r_{t-1}})\\ &= \mathbf{r_{t}} - <\mathbf{r_{t}}, \frac{\mathbf{r_{t-1}}}{||\mathbf{r_{t}}||}> \dfrac{\mathbf{r_{t-1}}}{||\mathbf{r_{t}}||}\\
    \label{eq:n-unit}
    \mathbf{1_{\mathbf{n}}} &= \dfrac{\mathbf{n}}{||\mathbf{n}||}
\end{align}

Where the vector projection of $\mathbf{r_{t}}$ onto $\mathbf{r_{t-1}}$ is represented by the red vector in Figure \ref{fig:pn-calc}. With the scalar $\dot{\Phi}_{\mathbf{\omega}}$ and the vector $\mathbf{1_{\mathbf{n}}}$, we can compute the desired acceleration as follows.

\begin{equation}
    \label{eq:my-des-acc}
    \mathbf{a}_{LOS'} = N V_c \dot{\Phi}_{\mathbf{\omega}} \mathbf{1_{\mathbf{n}}}
\end{equation}

In application on a quadrotor, in this work, the acceleration vector is fed into a velocity controller by integration of the command, which submits a roll, pitch, yawrate, thrust command to the internal autopilot controllers. Therefore, rather than adjusting heading to satisfy lateral accelerations, the application of TPN in this work relies on roll angle control. This more direct method of achieving lateral accelerations (that does not require forward velocity) is not possible on a missile or fixed-wing aircraft.

\subsection{Proportional Navigation with Heading Control}
\label{subsec:pn_with_hc}

The TPN algorithm presented above maintains the integrity of the algorithm commonly presented in missile guidance literature, but applies the control command more directly by controlling the roll angle of the UAV. During a missile's flight, the vehicle fulfills desired acceleration commands by flying in an arc, gradually changing its heading by relying on forward motion and the use of thrust vectoring or control surfaces. A quadrotor, however, has direct control over its heading by applying yaw-rate control. In this section, we describe an algorithm that uses PN acceleration in all axes but the lateral axis, and instead controls the heading to achieve lateral acceleration. Since it does not utilize PN in the lateral axis, we do not assign the label of ``true".

We define an inertial reference frame at the center of the UAV, with $x$ pointing forward, $y$ pointing to the left, and $z$ point upward. The acceleration along the $x$ and $z$ axes are simply taken from Equation \ref{eq:my-des-acc} as the corresponding components:

\begin{align}
    a_{x} &= \mathbf{a}_{LOS'} \bigcdot \mathbf{1}_x\\
    a_{z} &= \mathbf{a}_{LOS'} \bigcdot \mathbf{1}_z
\end{align}

The heading control composes of a commanded yaw-rate, which includes the computation of the heading:

\begin{align}
    \dot{yaw} &= K_{P,yaw} heading\\
    &= K_{P,yaw} \arctan{ \dfrac{\mathbf{r_t} \bigcdot \mathbf{1}_y}{\mathbf{r_t} \bigcdot \mathbf{1}_x} }
\end{align}

Where $K_{P,yaw}$ is a tuned proportional gain and $\mathbf{r_t}$ is the current LOS vector.

\subsection{Hybrid TPN-Heading Control}
\label{subsec:hybrid-tpn-hc}

There are potential benefits to both methods presented in Sections \ref{subsec:tpn-alg} and \ref{subsec:pn_with_hc}. TPN specifically applied to quadrotors via roll angle control might yield quicker reaction time for a moving object. PN while keeping the target centered in the frame ensures that the target is not lost from frame; otherwise, in a full system, the pursuing UAV would have to return to a search pattern. The goal of the hybrid algorithm is to capture the advantages of both methods.

This method switches between the two modes, \textit{PN} and \textit{Heading}. The transition between them simply relies on a tuned threshold $k_{heading}$ on the heading towards the target.

If $|heading| < k_{heading}$, enter state \textit{PN}:

\begin{align}
    a_{x} &= \mathbf{a}_{LOS'} \bigcdot \mathbf{1}_x\\
    a_{y} &= \mathbf{a}_{LOS'} \bigcdot \mathbf{1}_y\\
    a_{z} &= \mathbf{a}_{LOS'} \bigcdot \mathbf{1}_z\\
    \dot{yaw} &= 0.2 K_{P,yaw} heading
\end{align}

If $|heading| \geq k_{heading}$, enter state \textit{Heading Control}:

\begin{align}
    a_{x} &= \mathbf{a}_{LOS'} \bigcdot \mathbf{1}_x\\
    a_{y} &= 0.2 \mathbf{a}_{LOS'} \bigcdot \mathbf{1}_y\\
    a_{z} &= 0\\
    \dot{yaw} &= K_{P,yaw} heading
\end{align}

Where $k_{heading}$ may be tuned depending on certain factors of the UAV system, including the camera field-of-view (FOV) or the maximum yaw-rate. Note that at all times, regardless of the heading, both $a_{y}$ and $\dot{yaw}$ are nonzero, and are instead suppressed with a factor less than 1. The factor of 0.2 was found empirically in this study to perform well and yield an appropriate influence of both acceleration and yaw-rate. Using this hybrid method, the UAV may potentially be able to react to changes in target motion while also keeping it in view.

\section{Trajectory Following}
\label{sec:trajectory-following}

\begin{figure}[h!]
    \centering
    \includegraphics[height=5cm]{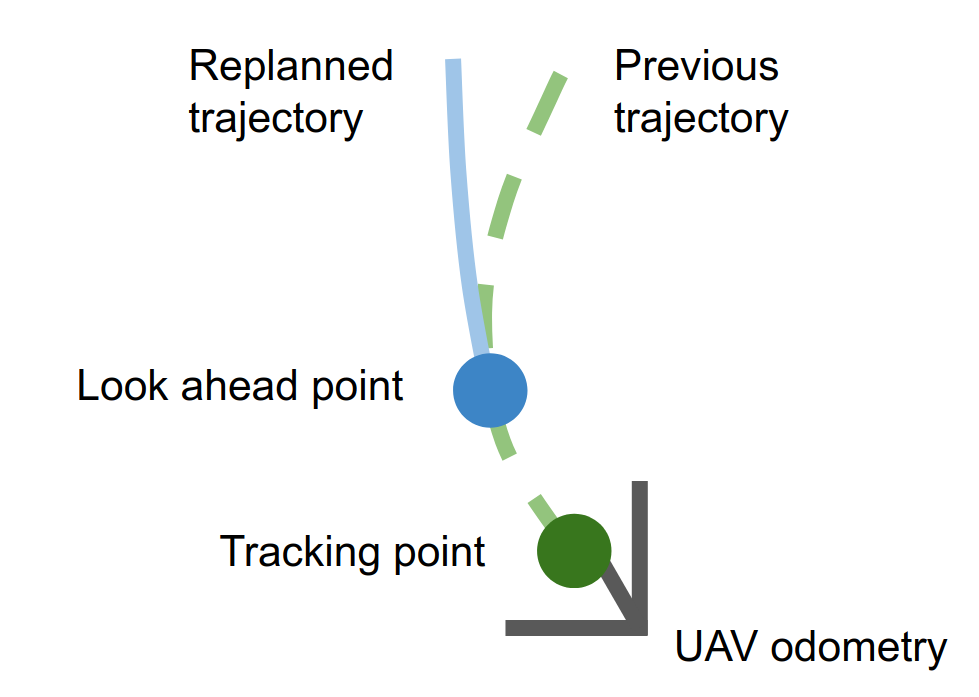}
    \caption{Diagram of trajectory replanning by stitching trajectory to the look ahead point. UAV closely follows the tracking point.}
    \label{fig:replanning}
\end{figure}

All trajectory following methods were implemented with some replanning rate at which updated trajectories are published. Replanning is constantly done from the look ahead point, which is maintained by the Trajectory Controller as described in Section \ref{subsec:system-control}.

\subsection{$LOS'$ Acceleration Trajectory Following}
\label{subsec:kin-traj}

These trajectories are formed by taking the desired acceleration command calculated with Equation \ref{eq:my-des-acc} and calculating position waypoints and velocities given the starting position set to the look ahead point. The calculations for the positions and velocities are done with the following kinematic equations, set in the UAV's inertial reference frame.

\begin{align}
    \mathbf{p}_t &= \mathbf{v}_0t + \frac{1}{2} \mathbf{a}_{LOS'} t^2\\
    \mathbf{v}_t &= \mathbf{v}_0 + \mathbf{a}_{LOS'} t , \, t=0,0.1,...,T
\end{align}

Where $T$ is the time length of each trajectory. As $T$ approaches $0$, this method becomes equivalent to commanding the desired $LOS'$ acceleration directly. The discretization of the timesteps is also a tunable parameter.

\subsection{Target Motion Forecasting}
\label{subsec:pred-traj}

In its simplest form, target motion forecasting involves estimating the target velocity in 3D, calculating the future location of the target with a straight-line path, and generating a collision course trajectory towards that point in space at a velocity which completes the trajectory at the specified time. This method makes three critical assumptions: (\textit{i}) the target has zero acceleration, (\textit{ii}) we can approximate time-to-collision by calculating the time along the current LOS vector, and (\textit{iii}) the forecasted target position will change slowly, so generating straight-path trajectories is sufficient to result in a final, smooth stitched trajectory. First, the LOS unit vectors at two times are calculated. These are used along with the depth estimation to find the target's velocity:

\begin{equation}
    \mathbf{v}_{target} = \dfrac{d_1(\mathbf{1_{LOS}}_1)-d_0(\mathbf{1_{LOS}}_0)}{t_1-t_0}
\end{equation}

$d_1$ and $d_0$, $\mathbf{1_{LOS}}_1$ and $\mathbf{1_{LOS}}_0$, and $t_1$ and $t_0$ are the estimated depth, calculated LOS unit vector, and time, at two timesteps. Once we have the velocity, we find the approximate time-to-collision along the current LOS vector by using the UAV's velocity component along the LOS:

\begin{equation}
    t_{collision} = \dfrac{d_1}{<\mathbf{v}_{UAV}, \mathbf{1_{LOS}}_1>}
\end{equation}

Where $\mathbf{v}_{UAV}$ is the current UAV velocity vector. Therefore, the approximate point in space to plan the trajectory to is as follows, where $\mathbf{p}_{collision}$ is in the UAV reference frame.

\begin{equation}
\label{eq:fortraj-p-collision}
    \mathbf{p}_{collision} = \mathbf{v}_{target}t_{collision} + d_1(\mathbf{1_{LOS}}_1)
\end{equation}

\section{System}
\label{sec:system}

\begin{figure}[h!]
    \centering
    \includegraphics[width=0.9\linewidth]{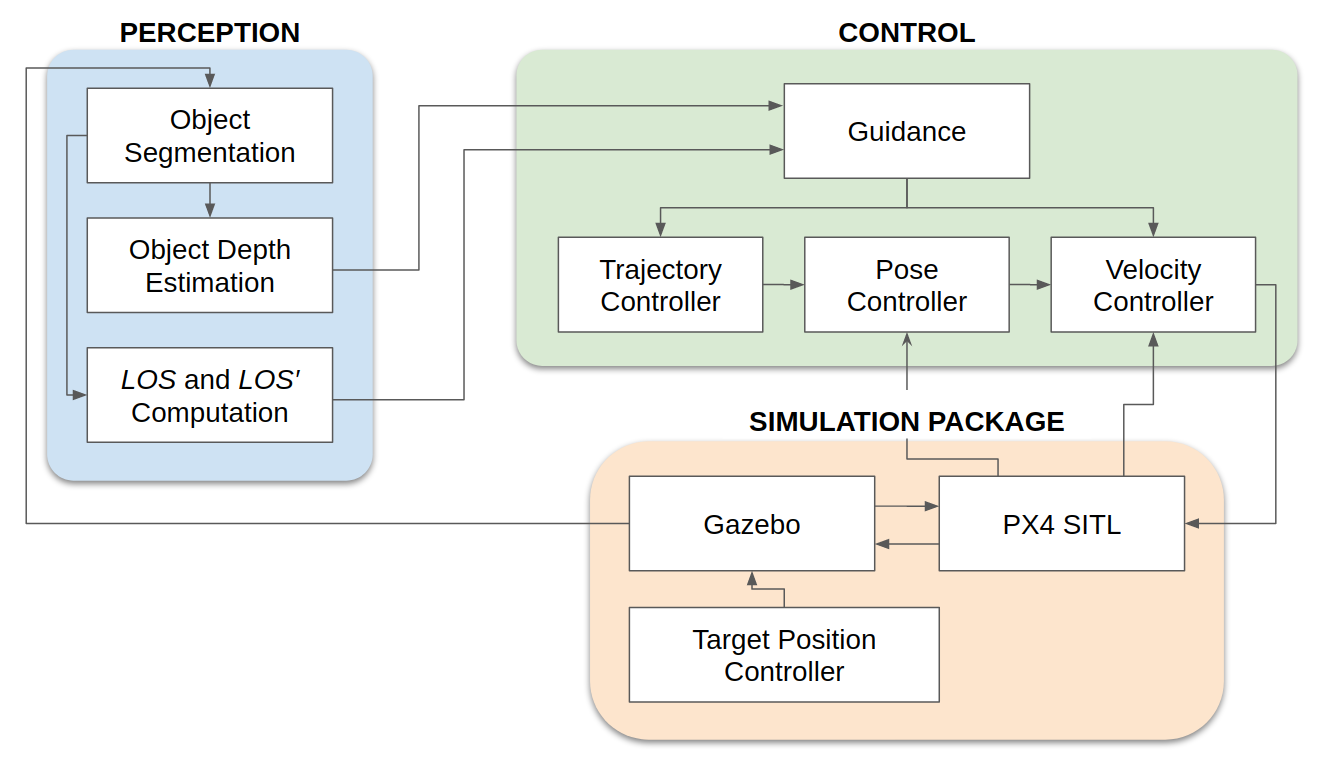}
    \caption{System diagram.}
    \label{fig:software-diagram}
\end{figure}

Figure \ref{fig:software-diagram} shows a diagram of the most important parts of the system, including perception, control, and the simulation software. Each block generally consists of one or two nodes in ROS (Robot Operation System), the framework used in this work.

\subsection{Perception}
\label{subsec:system-perception}

\subsubsection{Object Segmentation}

\begin{figure}[h!]
\centering
\subcaptionbox{\label{fig:target-frame}RGB image frame with target in view during pursuit.}{\includegraphics[height=5.7cm]{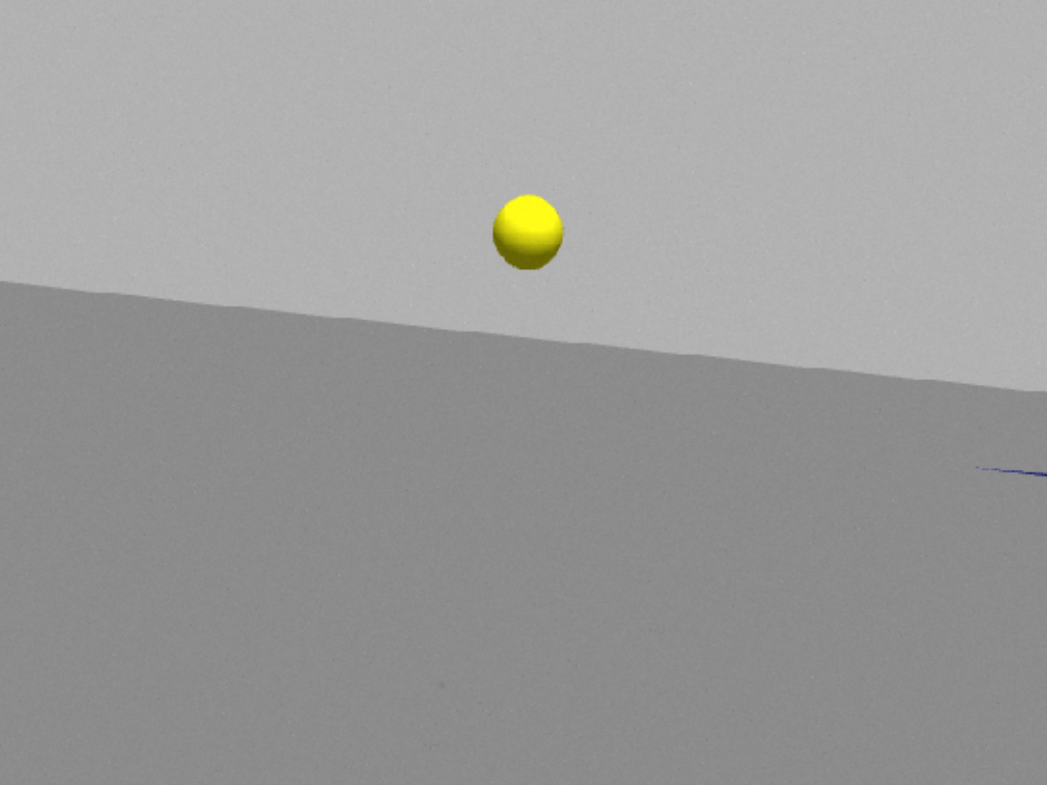}}\hfill
\subcaptionbox{\label{fig:target-segmented}Binary segmentation image with target highlighted.}{\includegraphics[height=5.7cm]{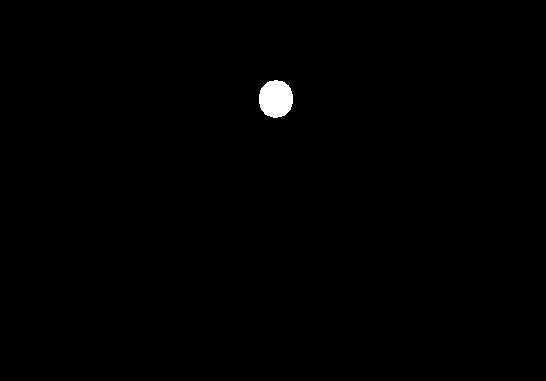}}\\
\caption{Input and output of Object Segmentation node.}
\label{fig:object-segmentation-inout}
\end{figure}

% \begin{figure}[h!]
%     \begin{minipage}{.5\textwidth}
%     \centering
%     \includegraphics[height=4cm]{images/image-target.png}
%     \caption{}
%     \label{fig:target-frame}
%     \end{minipage}%
%     \centering
%     \begin{minipage}{.5\textwidth}
%     \centering
%     \includegraphics[height=4cm]{images/image-segmentation.png}
%     \caption{}
%     \label{fig:target-segmented}
%     \end{minipage}
% \end{figure}

Since the focus for this chapter was on the comparison of guidance methods, the perception challenge was simplified. The simulation does not include rich background visuals, there is no dropout in the camera feed, and the object was reduced from a potential UAV shape to a yellow sphere. The segmentation node composes of color thresholding that creates a binary segmentation image from each camera frame in real-time (30Hz), as seen in Figure \ref{fig:target-segmented}. In addition, this node publishes the centroid $(C_x,C_y)$ of the detected object by using the image moments.

\begin{equation}
    \begin{bmatrix}
        C_x \\ C_y
    \end{bmatrix}
    =
    \begin{bmatrix}[2.0]
        \dfrac{M_{10}}{M_{00}} \\ \dfrac{M_{01}}{M_{00}}
    \end{bmatrix}
\end{equation}

\subsubsection{Object Depth Estimation}

\begin{figure}[h!]
    \begin{minipage}{.5\textwidth}
    \centering
    \includegraphics[width=0.9\textwidth]{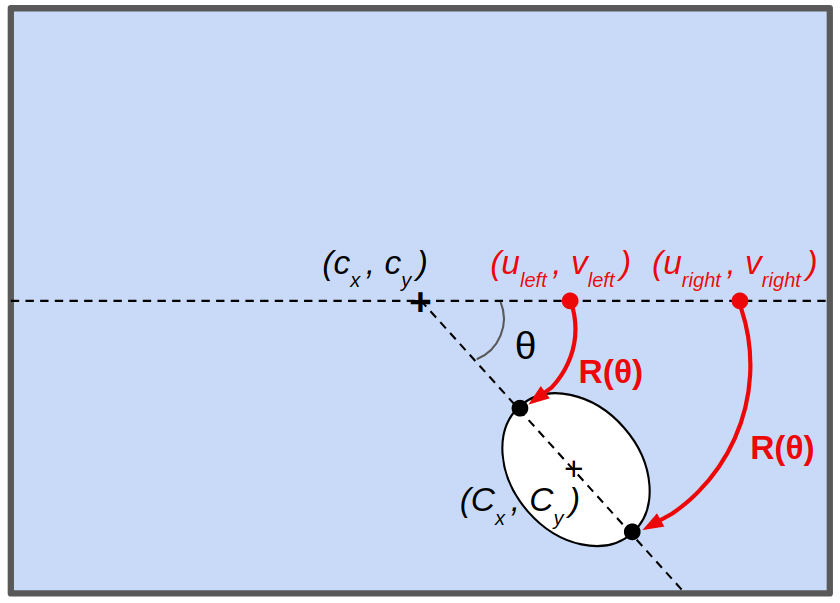}
    \caption{Example image frame with detected object's centroid and part of depth estimation calculations. $(u_{left},v_{left})$ and $(u_{right},v_{right})$ points along image horizontal corresponding to target edges are shown before (red) and after (black) rotation back into the original segmented coordinates. Note that because of the wide field of view, images around the edges of the image frame can get warped even though there is no distortion in this camera model.}
    \label{fig:depth-est-image}
    \end{minipage}%
    \centering
    \begin{minipage}{.5\textwidth}
    \centering
    \includegraphics[width=0.9\linewidth]{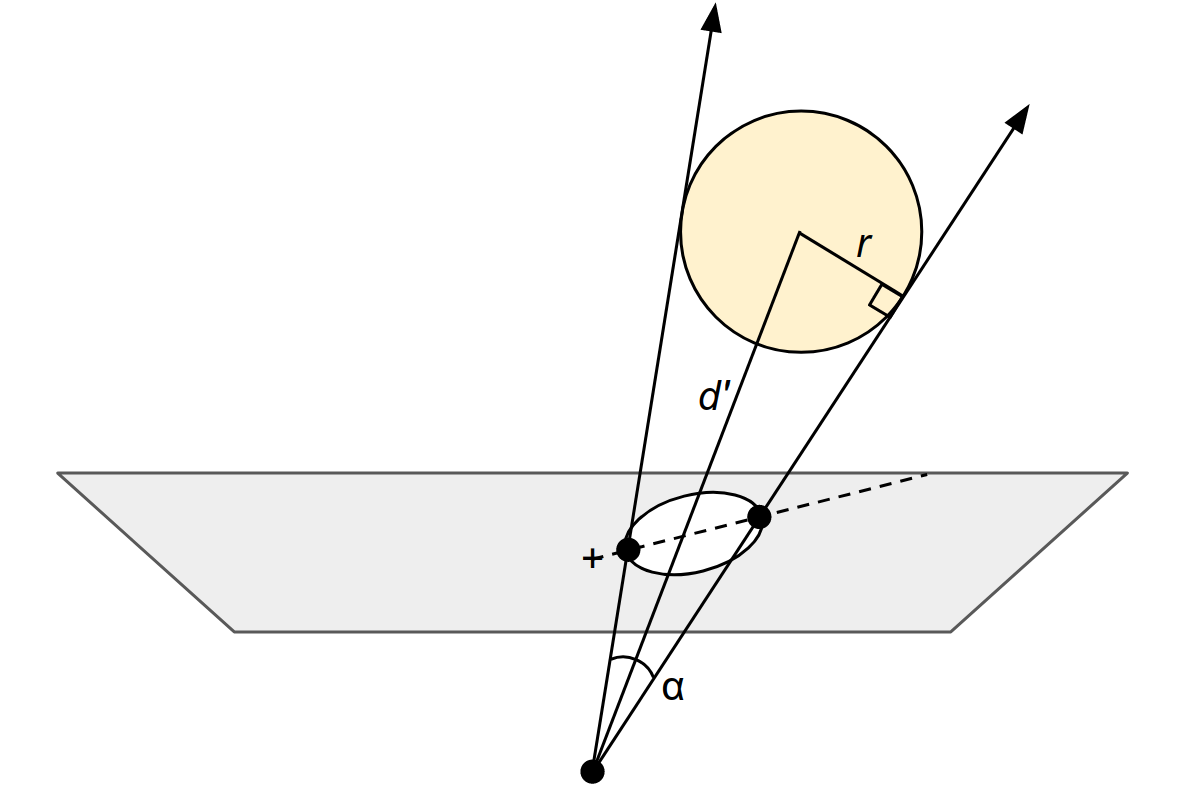}
    \caption{Top-down diagram showing calculation of estimated distance to object center $d'$ with 3D projected rays to target edges.}
    \label{fig:depth-est-calc}
    \end{minipage}
\end{figure}

Depth estimation with monocular camera images relies on prior knowledge of the target size as well as the assumption that the target is spherical (which is correct in this simulation, but extendable to real-life targets with approximate knowledge of target or aircraft type). The estimate takes into account the camera intrinsic parameters and projects the 2D information (pixels) to 3D (rays starting at the camera center) to extract the depth.

Pixel coordinates at two points in the image need to be located which form a 2D line in the image that projects to a 3D line along the diameter of the object. The easiest found such line is the line through the image principal point that passes through the centroid, since the target is a sphere. This line is the dotted line passing through the target in Figure \ref{fig:depth-est-image}. We must first find the pixel coordinates along this line that correspond to the object edges in the image.

First, we find the rotation angle formed by the centroid coordinates with respect to the image horizontal, labeled $\theta$ in Figure \ref{fig:depth-est-image}.

\begin{equation}
    \theta = \arctan \dfrac{C_y-c_y}{C_x-c_x}
\end{equation}

With this angle we can use a rotation matrix to rotate the segmented object onto the $y=c_y$ line. Note that we use $-\theta$ to achieve the correct rotation, and that we first compute new pixel coordinates $(u_i,v_i)$ relative to the image center, from the original coordinates $(x_i,y_i)$.

\begin{align}
    \begin{bmatrix}
    u_{i} \\ v_{i}
    \end{bmatrix}
    &=
    \begin{bmatrix}
    x_i - c_x \\ y_i - c_y
    \end{bmatrix}
    , \; i=1,...,N \\
    \begin{bmatrix}
    u_{i,R} \\ v_{i,R}
    \end{bmatrix}
    &=
    \begin{bmatrix}
    cos(-\theta) && -sin(-\theta)\\
    sin(-\theta) && cos(-\theta)
    \end{bmatrix}
    \begin{bmatrix}
    u_{i} \\ v_{i}
    \end{bmatrix}
    , \; i=1,...,N
\end{align}

Where N is the total number of segmented pixels in the binary segmentation image. Once the object is rotated onto the image horizontal, simple \texttt{min} and \texttt{max} operations yield the extreme pixel coordinates on the left and right of the target (or right and left, if $C_x<c_x$).

\begin{align}
    \begin{bmatrix}
    u_{left} \\ v_{left}
    \end{bmatrix}
    &=
    \begin{bmatrix}
    min_i(u_i) \\ c_y
    \end{bmatrix}
    , \; i=1,...,N\\
    \begin{bmatrix}
    u_{right} \\ v_{right}
    \end{bmatrix}
    &=
    \begin{bmatrix}
    max_i(u_i) \\ c_y
    \end{bmatrix}
    , \; i=1,...,N
\end{align}

These two points are shown in red in Figure \ref{fig:depth-est-image}. We apply the inverse rotation and translation to get the original coordinates of these two identified pixel coordinates.

\begin{align}
    \begin{bmatrix}
    u_{left,R^{-1}} \\ v_{left,R^{-1}}
    \end{bmatrix}
    &=
    \begin{bmatrix}
    cos(\theta) && -sin(\theta)\\
    sin(\theta) && cos(\theta)
    \end{bmatrix}
    \begin{bmatrix}
    u_{left} \\ v_{left}
    \end{bmatrix}
    +
    \begin{bmatrix}
    c_x \\ c_y
    \end{bmatrix}\\
    \begin{bmatrix}
    u_{right,R^{-1}} \\ v_{right,R^{-1}}
    \end{bmatrix}
    &=
    \begin{bmatrix}
    cos(\theta) && -sin(\theta)\\
    sin(\theta) && cos(\theta)
    \end{bmatrix}
    \begin{bmatrix}
    u_{right} \\ v_{right}
    \end{bmatrix}
    +
    \begin{bmatrix}
    c_x \\ c_y
    \end{bmatrix}
\end{align}

These two pixels are projected to 3D rays similar to the LOS ray calculation in Equation \ref{eq:los}. They can also be seen in a similar top-down view in Figure \ref{fig:depth-est-calc}. Once the two rays are found, the depth (to the nearest point on the object) is as follows:

\begin{equation}
    d = \dfrac{w/2}{\sin{(\alpha/2)}} - \dfrac{w}{2}
\end{equation}

% TODO include plot of estimated vs actual depth as object moves from very close to far away

\subsubsection{$LOS$ and $LOS'$ Computation}

The calculation of $\mathbf{r}_t$, the LOS vector, can be found in Equation \ref{eq:los}. The calculations of $\dot{\Phi}_{\mathbf{\omega}}$ and $\mathbf{1}_{\mathbf{n}}$, the scaling and directional components of LOS', can be found in Equations \ref{eq:phi-omega-dot} and \ref{eq:n-unit}. Before being input to the Guidance node, the quantities undergo smoothing with a flat moving average filter. The filtered values are used for trajectory-based guidance, while LOS-based guidance uses the raw values.

\subsection{Control}
\label{subsec:system-control}

\subsubsection{Guidance}

The Guidance node either executes LOS-based or trajectory-based guidance. The LOS-based guidance, as described in Section \ref{sec:los-guidance}, utilizes the LOS and LOS' computation from the Perception block. In this case, the node outputs acceleration commands that are satisfied by the Velocity Controller. Trajectory-based guidance utilizes both the LOS information as well as the depth estimate to generate 3D trajectories towards the target's current state or forecasted motion. Here, the node outputs a trajectory composing of waypoints and corresponding speeds which is accepted by the Trajectory Controller. Every method is initialized with two seconds of simple LOS guidance, where the UAV accepts velocity commands directly along the current LOS vector.

\subsubsection{Trajectory, Pose, and Velocity Controllers}

The Trajectory Controller accepts a trajectory in the form of a list of waypoints, each with a corresponding $(x,y,z)$ position, yaw, and speed (scalar). It outputs a tracking point, which the Pose Controller takes as input for tracking the trajectory, and a look ahead point, which is used as the replanning point. The configurable look ahead time sets how far ahead the look ahead point is from the tracking point, and is approximated to be $1/f + \Delta buffer$, where $f$ is the replanning frequency and $\Delta buffer$ is a buffer in case there is some lag in the system. The Pose Controller composes of PID controllers with the tracking point from the Trajectory Controller as the reference and the odometry from the PX4 SITL interface as the actual. This outputs a velocity reference (and a velocity feedforward term from the Trajectory Controller) to the Velocity Controller, which also accepts the odometry from the PX4 SITL interface as the actual. The Velocity Controller also uses PID controllers, and outputs a roll, pitch, yawrate, thrust commands to the PX4 SITL interface.

\subsection{Simulation Package}
\label{subsec:system-simulation}

\subsubsection{Gazebo and PX4 SITL}

\begin{figure}[h!]
    \centering
    \includegraphics[width=0.7\textwidth]{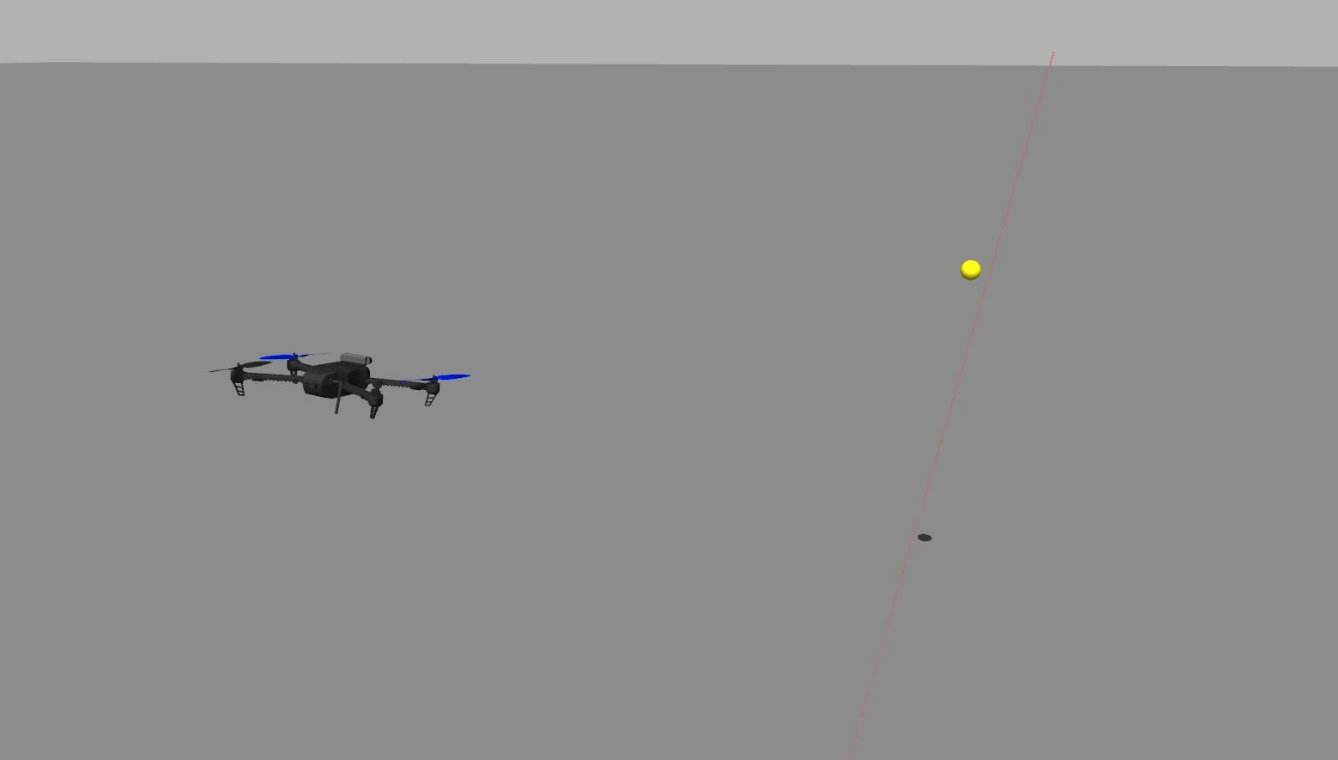}
    \caption{Gazebo simulation world, with UAV model and target in view.}
    \label{fig:gazebo-screenshot}
\end{figure}

Simulation was used to develop, deploy, and evaluate many UAV guidance algorithms quickly. In this environment, the behavior of the UAV and the target was easily modified and could be strained without the safety risk involved in real-world testing. Gazebo \cite{koenig2004design}, the simulator of choice, is commonly used for robotics testing and simulates robot dynamics with the ODE physics engine. This was paired with PX4 autopilot \cite{lorenz_meier_2020_3819570} software-in-the-loop (SITL) for control of the quadrotor. An example of the simulated world can be seen in Figure \ref{fig:gazebo-screenshot}. Obstacles, a visually rich backdrop, and other factors were eliminated in the simulation setup to reduce the impact of external factors on the evaluation of the guidance algorithms. When deployed on a real-world UAS, the algorithms here can be merged within a larger autonomy architecture including robust detection and tracking, as well as obstacle avoidance and others. The quadrotor is outfitted with a RGB camera that models a realistic vision sensor.

\subsubsection{Target Position Controller}

\begin{figure}[H]
\centering
\subcaptionbox{\label{fig:traj-stationary}Straight target path.}{\includegraphics[height=3.5cm]{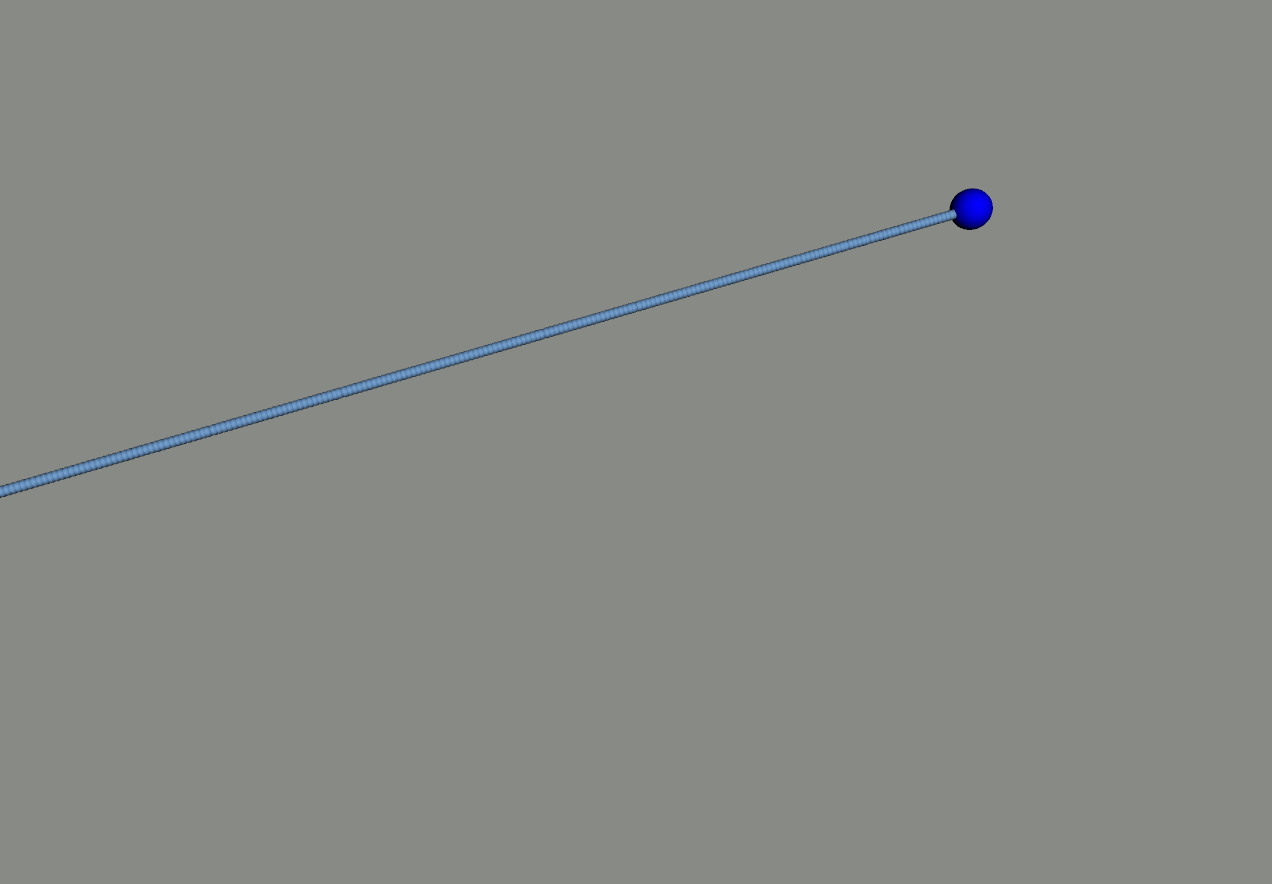}}\hfill
\subcaptionbox{\label{fig:traj-fig8}Figure-8 target path.}{\includegraphics[height=3.5cm]{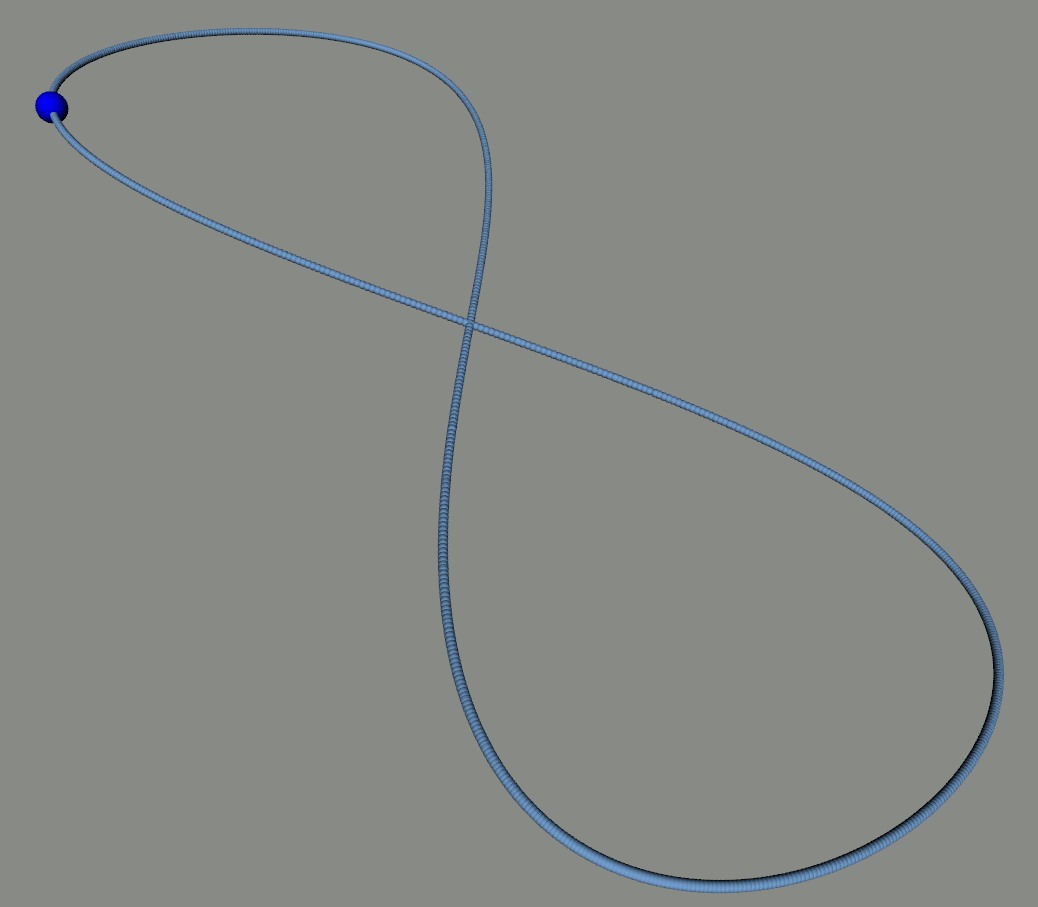}}\hfill
\subcaptionbox{\label{fig:traj-random}Knot target path.}{\includegraphics[height=3.5cm]{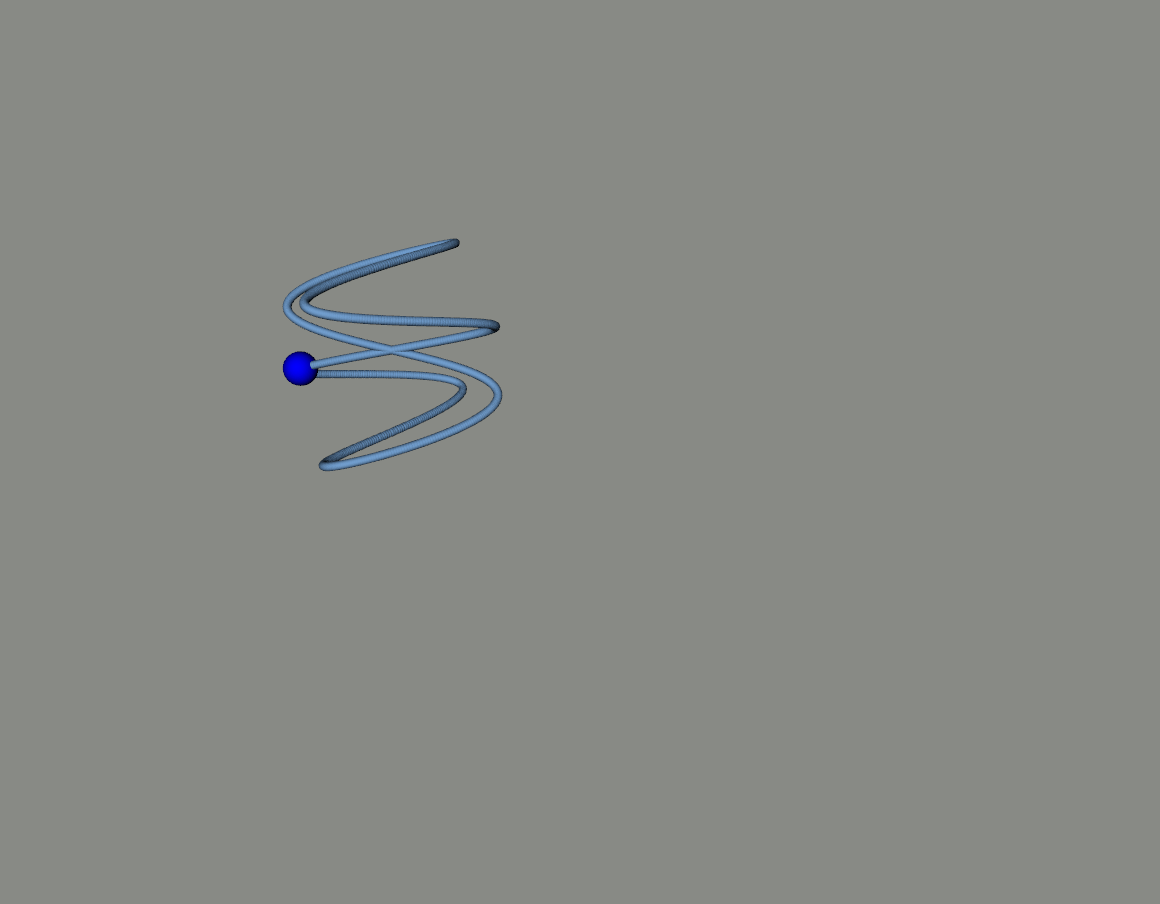}}\\
\caption{Target path library taken from similar perspectives. Dark blue sphere is the target's current position. Light blue sequence is the traced path. (a) straight path with random slope; (b) figure-8 path with random tilt, of size 10m$\times$6m; (c) knot path with random center point, of size 2m$\times$2m$\times$2m.}
\label{fig:target-trajs}
\end{figure}

% \begin{figure*}[ht!]
%   \subfloat[\label{fig:traj-stationary}]{%
%       \includegraphics[trim=20 40 50 30,clip, height=3.5cm]{figs/rviz-sl.png}}
% \hspace{\fill}
%   \subfloat[\label{fig:traj-random} ]{%
%       \includegraphics[trim=10 40 10 30,clip, height=3.5cm]{images/traj-random.png}}
% \hspace{\fill}
%   \subfloat[\label{fig:traj-fig8}]{%
%       \includegraphics[trim=30 40 30 30,clip, height=3.5cm]{images/traj-fig8.png}}\\
% \caption{\label{fig:target-trajs}Target trajectory library taken from similar perspectives. Dark blue sphere is the target's current position. Light blue sequence is the traced path. (a) straight path with random slope; (b) knot shape with random center point, within 2m$\times$2m$\times$2m; (c) figure-8 trajectory with random tilt.}
% \end{figure*}

A library of target paths was used to strain the UAV's capability of intercepting the target. The first, simplest target motion is a straight path with constant velocity, which may mimic an autonomous aircraft on a search pattern or a fixed-wing plane. The starting position places the target on either side of the UAV's FOV, and the path crosses in front of the UAV with some random variation in slope. The third trajectory is a figure-8 with a randomized 3D tilt, similar to an evasive maneuver a small aircraft may take. These trajectories can be seen in Figure \ref{fig:target-trajs}. The third trajectory composes of a knot shape filling a 2m$\times$2m$\times$2m space at a random location within a 10m$\times$20m$\times$10m area in front of the UAV. This more rapid movement back and forth is similar to a multi-rotor hovering in a changing wind field.

\section{Results}
\label{sec:results}

In this section, we evaluate the five guidance algorithms described in the sections prior (Sections \ref{subsec:tpn-alg}, \ref{subsec:pn_with_hc}, \ref{subsec:hybrid-tpn-hc}, \ref{subsec:kin-traj}, \ref{subsec:pred-traj}). An experiment configuration composed of a selection of one parameter from each of the following categories. Each configuration underwent 50 trials.

\begin{itemize}
    \item [] \textbf{UAV Guidance:} True Proportional Navigation (TPN), Proportional Navigation with Heading Control (PN-Heading), Hybrid True Proportional Navigation with Heading Control (Hybrid TPN-Heading), LOS' Trajectory, Forecasting Trajectory
    \item [] \textbf{UAV speed [m/s]:} 2.0, 3.0, 4.0, 5.0
    \item [] \textbf{Target path:} Straight, Figure-8, Knot
    \item [] \textbf{Target speed [\% of UAV speed]:} 25\%, 50\%, 75\%, 100\%
\end{itemize}

In the case of sinusoidal target paths (figure-8 and knot), the target speed was set by determining the length of the path and dividing by the desired speed to calculate the period of the sinusoids.

The primary metric for comparing different methods is the first-pass hit rate, presented in Section \ref{subsec:first-pass-hit-rates}. All of the following conditions must be met for a trial to be considered a successful first-pass hit on the target.

\begin{enumerate}
    \item UAV is within 0.5m of the target (measured from the closest point on the target to the center of the UAV).
    \item Duration of pursuit is less than 20s.
    \item UAV stays within a 35m$\times$100m$\times$40m area surrounding the target.
    \item Target is not outside the UAV camera's FOV for more than 3s.
\end{enumerate}

These conditions were specified after consideration of the simulation scene, UAV model, and target model. The target is a yellow sphere of diameter 1m. The RGB camera has a horizontal FOV of 105$\degree$ with a size 680$\times$480. In lieu of a gimbal camera in simulation, the UAV model had a rigidly fixed, angled camera that was adjusted for each speed, to compensate for the steady state pitch down when flying forward at high speeds.

% [maybe not; probably unnecessary] TODO picture of camera tilted upwards for each speed

The next metric presented is the mean pursuit durations, in Section \ref{subsec:pursuit-duration}. These were computed by taking the mean of the time-to-hit measurements over each of (a) target speed, and (b) UAV velocity, respectively. This was done using only the successful hits during each experiment.

% assumptions?

\subsection{First-Pass Hit Rates}
\label{subsec:first-pass-hit-rates}

Datapoints in the following heatmaps that are inside a black square represent experiments that were not able to complete more than 95\% of the trials without crashing due to UAV instability with the particular experiment configuration.

\begin{figure}[H]
\centering
\subcaptionbox{\label{tpn-hit-rate:a}Straight target path.}{\includegraphics[width=.32\linewidth]{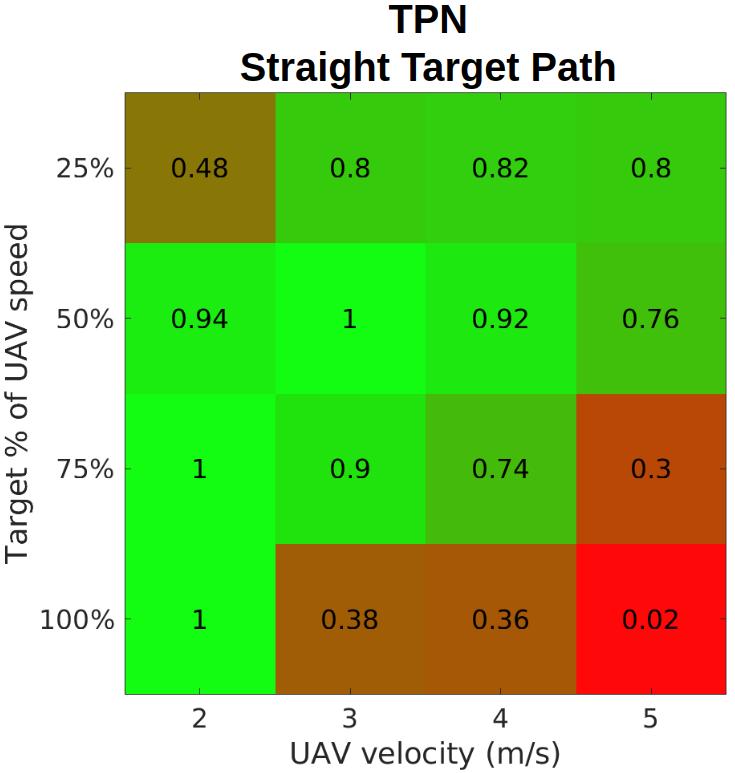}}\hfill
\subcaptionbox{\label{tpn-hit-rate:b}Figure-8 target path.}{\includegraphics[width=.32\linewidth]{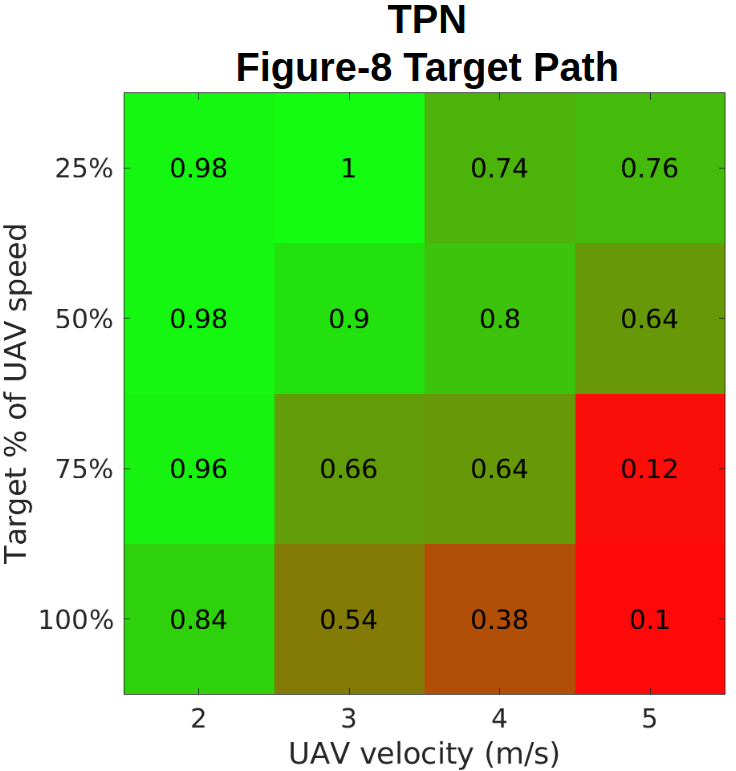}}\hfill
\subcaptionbox{\label{tpn-hit-rate:c}Knot target path.}{\includegraphics[width=.32\linewidth]{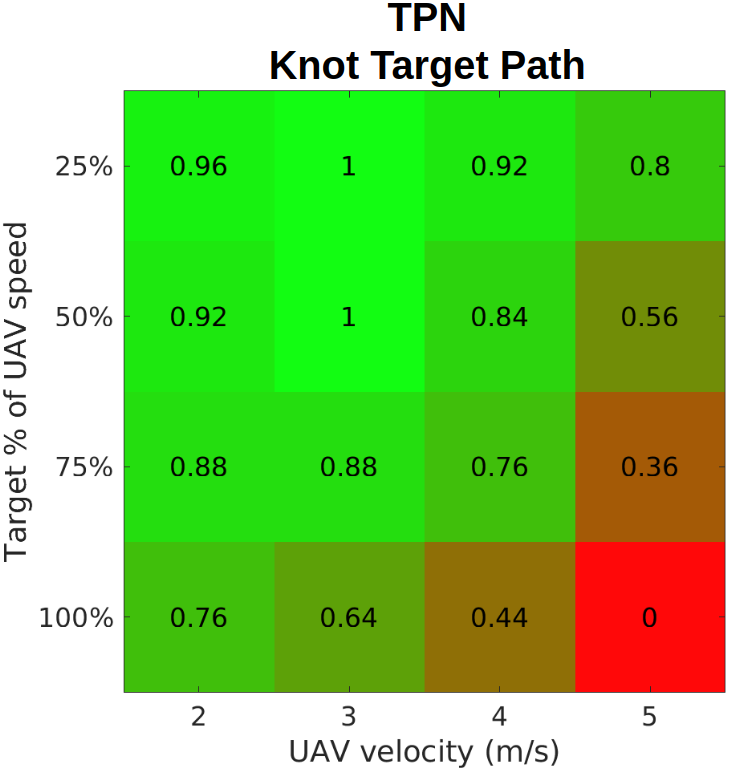}}\\
\caption{True Proportional Navigation hit rate across three target paths.}
\label{fig:tpn-hit-rate}
\end{figure}

\begin{figure}[h!]
\centering
\subcaptionbox{\label{pnhc-hit-rate:a}Straight target path.}{\includegraphics[width=.32\linewidth]{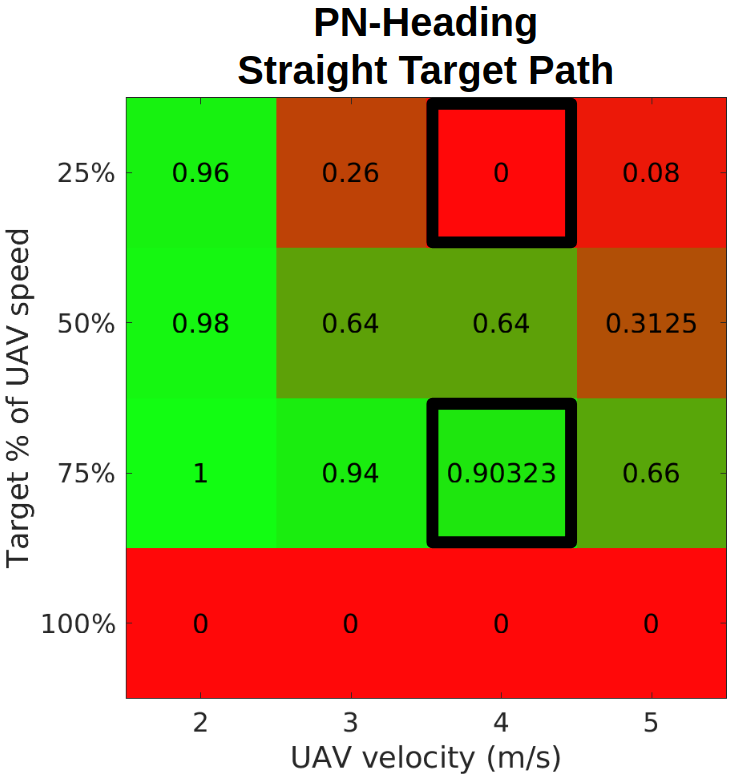}}\hfill
\subcaptionbox{\label{pnhc-hit-rate:b}Figure-8 target path.}{\includegraphics[width=.32\linewidth]{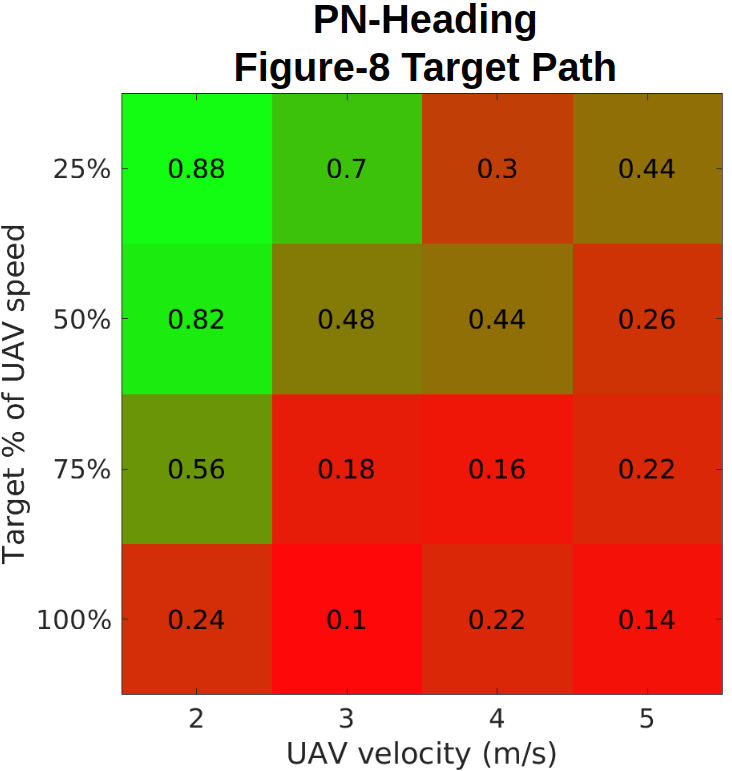}}\hfill
\subcaptionbox{\label{pnhc-hit-rate:c}Knot target path.}{\includegraphics[width=.32\linewidth]{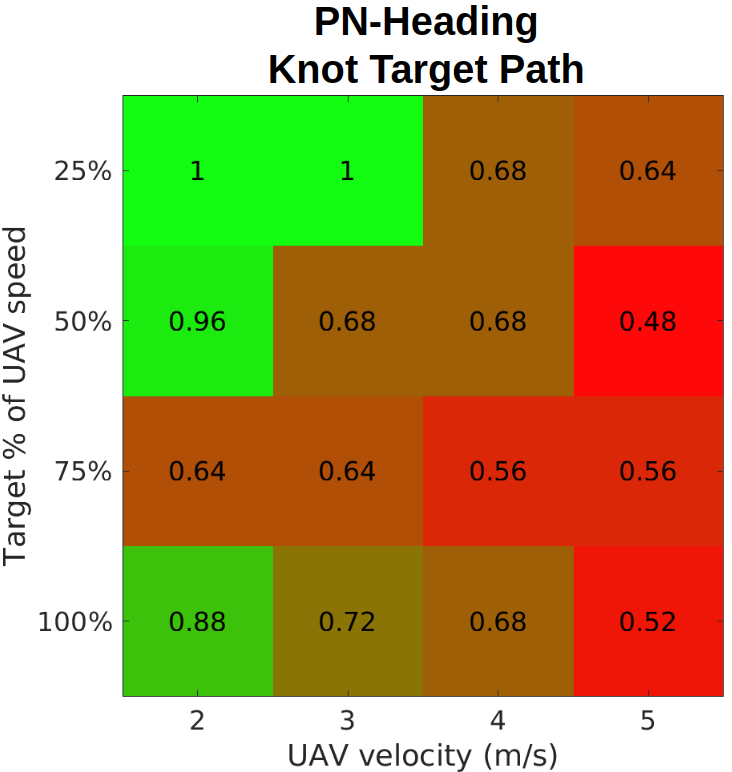}}\\
\caption{Proportional Navigation with Heading Control hit rate across three target paths.}
\label{fig:pnhc-hit-rate}
\end{figure}

\begin{figure}[h!]
\centering
\subcaptionbox{\label{hybrid-hit-rate:a}Straight target path.}{\includegraphics[width=.32\linewidth]{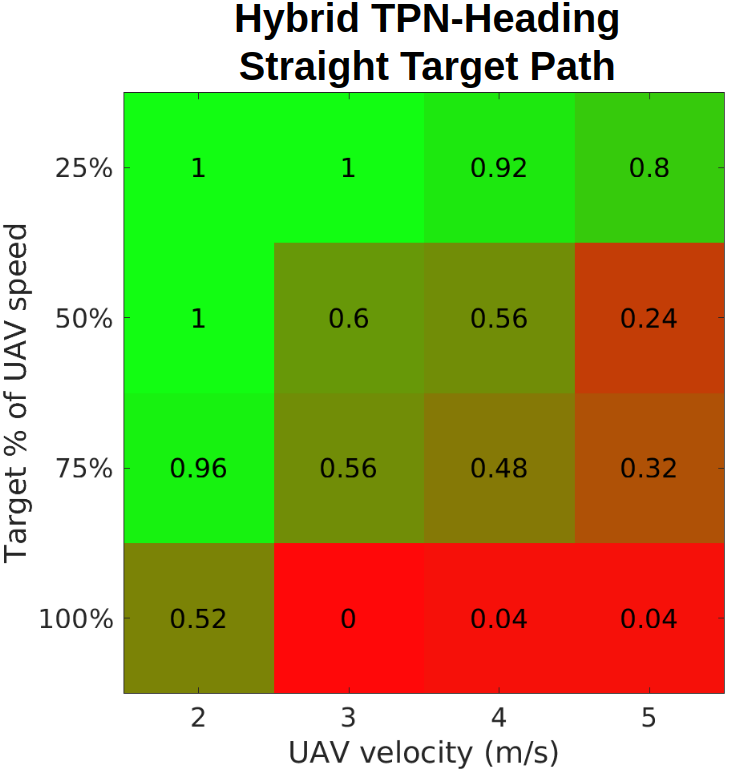}}\hfill
\subcaptionbox{\label{hybrid-hit-rate:b}Figure-8 target path.}{\includegraphics[width=.32\linewidth]{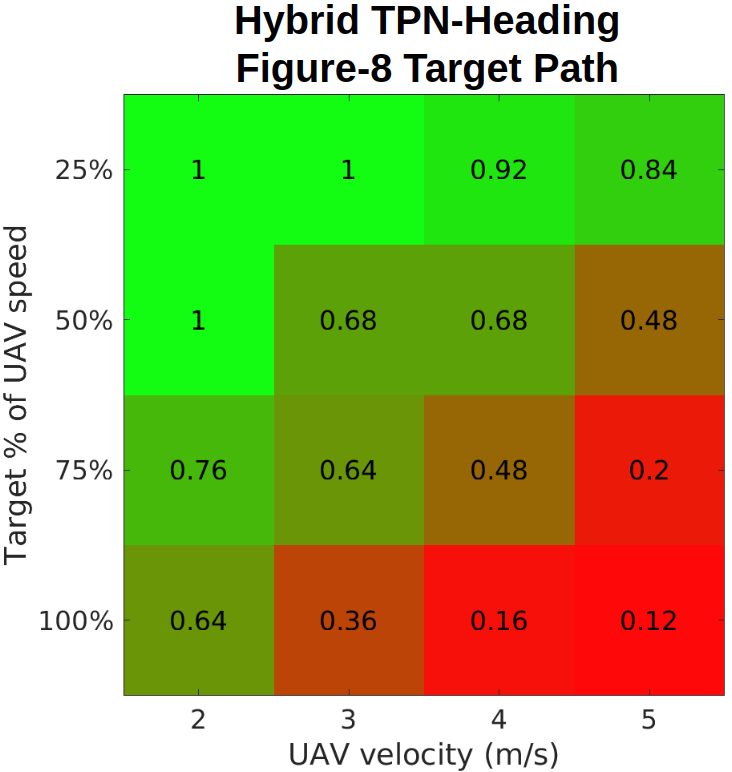}}\hfill
\subcaptionbox{\label{hybrid-hit-rate:c}Knot target path.}{\includegraphics[width=.32\linewidth]{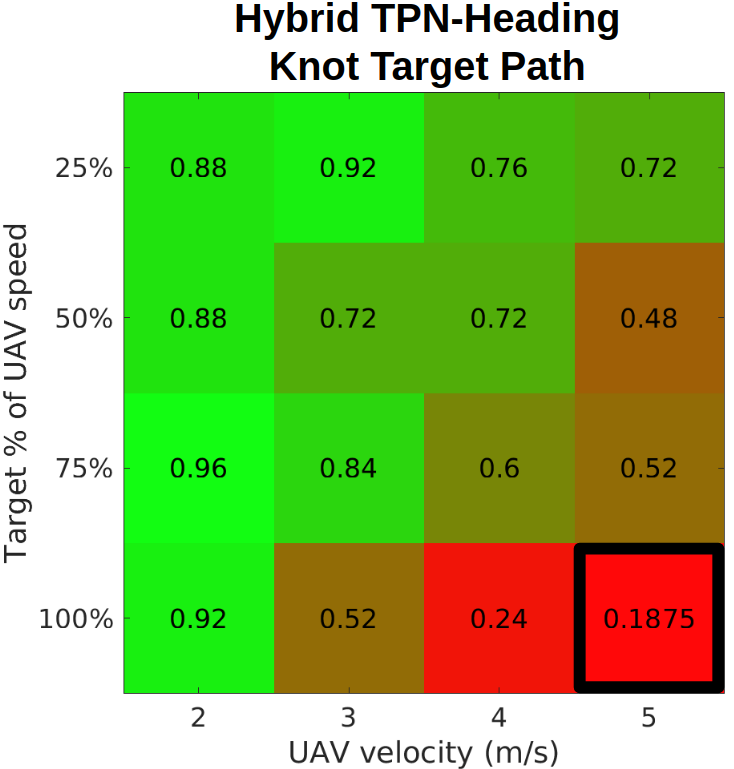}}\\
\caption{Hybrid True Proportional Navigation-Heading Control hit rate across three target paths.}
\label{fig:hybrid-hit-rate}
\end{figure}

\begin{figure}[h!]
\centering
\subcaptionbox{\label{lostraj-hit-rate:a}Straight target path.}{\includegraphics[width=.32\linewidth]{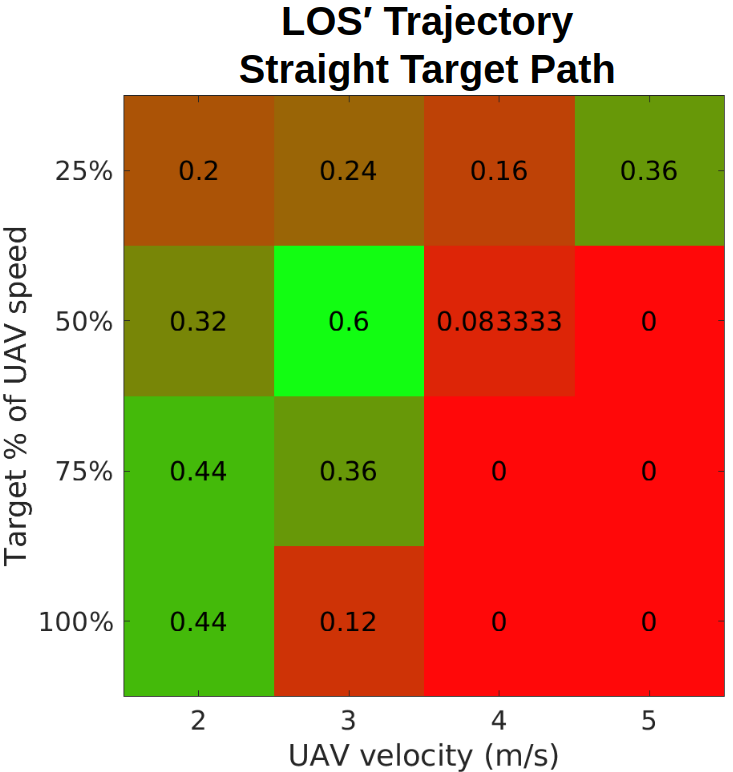}}\hfill
\subcaptionbox{\label{lostraj-hit-rate:b}Figure-8 target path.}{\includegraphics[width=.32\linewidth]{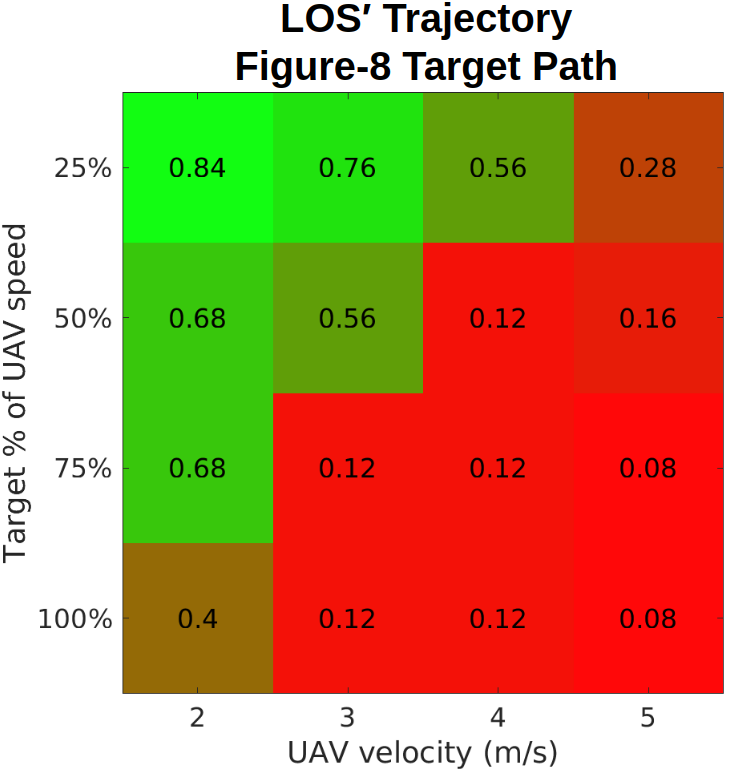}}\hfill
\subcaptionbox{\label{lostraj-hit-rate:c}Knot target path.}{\includegraphics[width=.32\linewidth]{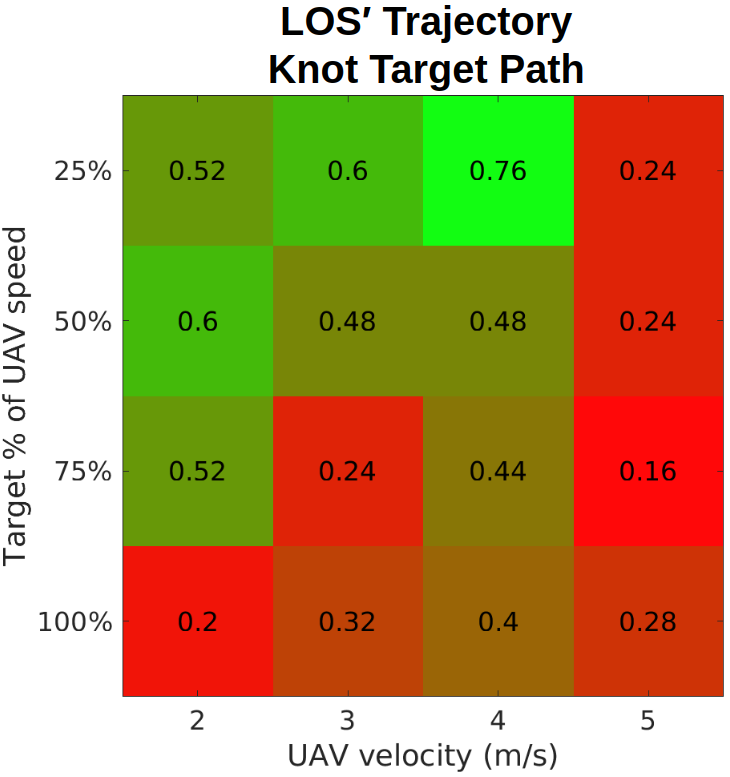}}\\
\caption{LOS' Trajectory hit rate across three target paths.}
\label{fig:lostraj-hit-rate}
\end{figure}

\begin{figure}[h!]
\centering
\subcaptionbox{\label{fortraj-hit-rate:a}Straight target path.}{\includegraphics[width=.32\linewidth]{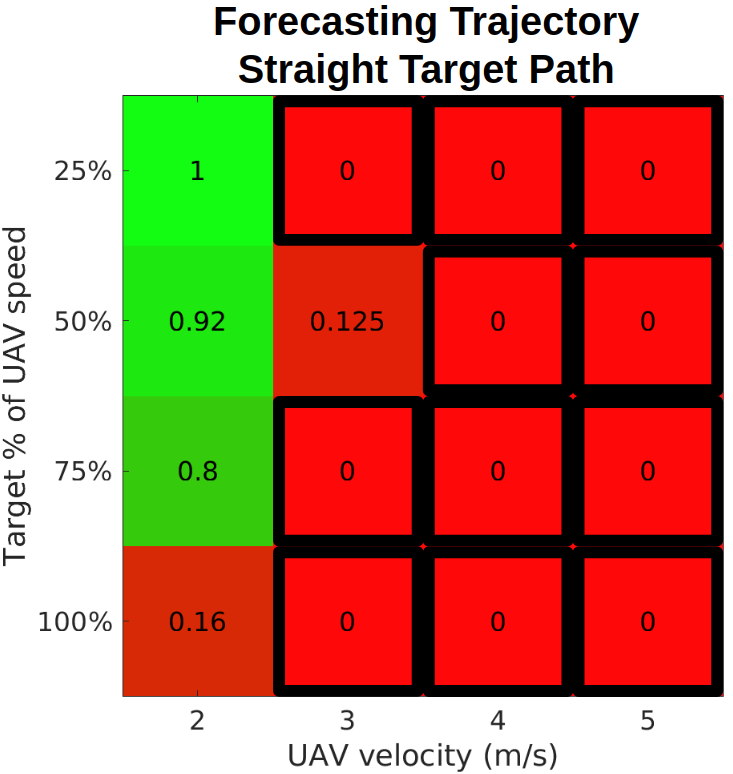}}\hfill
\subcaptionbox{\label{fortraj-hit-rate:b}Figure-8 target path.}{\includegraphics[width=.32\linewidth]{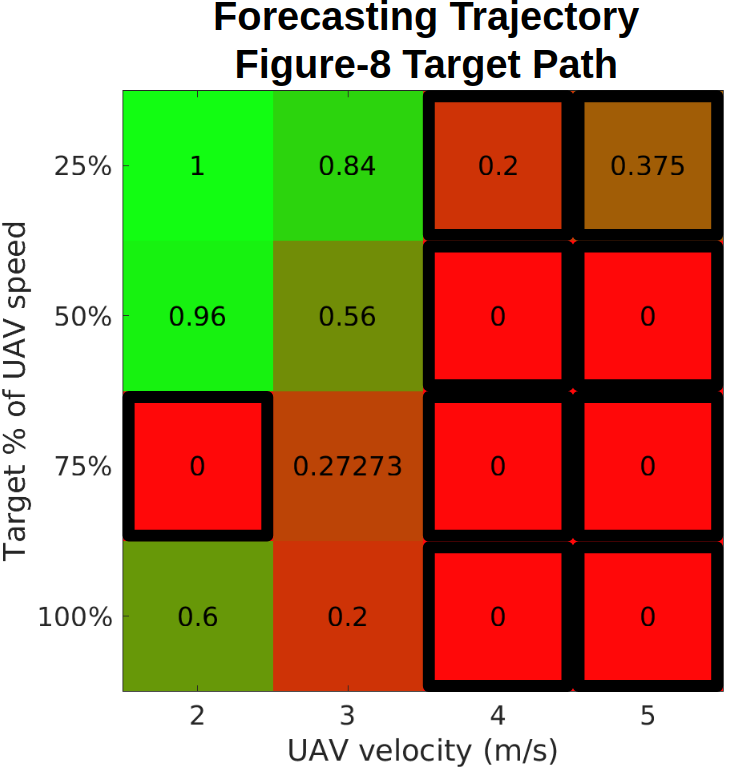}}\hfill
\subcaptionbox{\label{fortraj-hit-rate:c}Knot target path.}{\includegraphics[width=.32\linewidth]{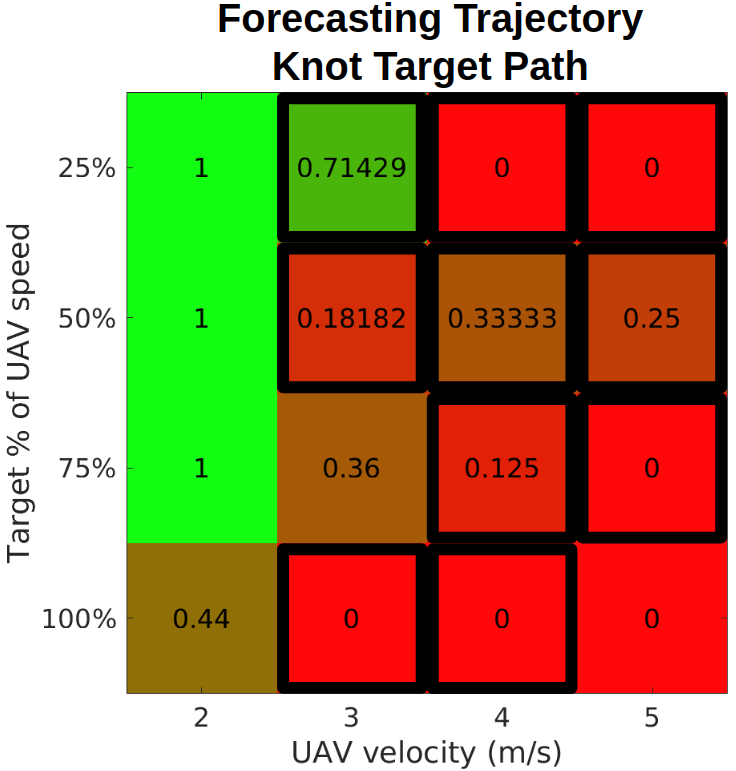}}\\
\caption{Forecasting Trajectory hit rate across three target paths.}
\label{fig:fortraj-hit-rate}
\end{figure}

TPN achieves the highest hit rate across almost all configurations compared to the other methods in both classes (LOS Guidance and Trajectory Following). Across all experiments there is a trend of lower UAV and target speeds resulting in higher hit rates, sometimes even of 1.0 (i.e., 50 of 50 trials result in a hit). Moving down and to the right within each subfigure presents results at higher target and UAV speeds. This can be seen as increasing the \textit{closing velocity}, which reduces the time that the UAV has to react to changes in target motion. As seen in the figures, this results in lower hit rates, which may be due to lag in the UAV controllers' ability to fulfill desired acceleration commands. The quadorotor achieves a lateral acceleration more directly than a fixed-wing craft by inducing a roll angle and thereby shifting a component of the thrust to this axis. However, the moment of the aircraft and the response time of the controller both contribute to lag in achieving the necessary roll angle. Therefore, as the time allowed for this control decreases, with the increase in closing velocity, it is more unlikely that the necessary lateral acceleration will be achieved. Similar logic applies to the thrust controller for achieving desired acceleration in the $z$ axis. It is possible that a more accurate controller might increase the hit rates at high closing velocities.

The Hybrid TPN-Heading method generally had higher hit rates than PN-Heading, but lower than TPN. However, Hybrid TPN-Heading consistently performed better than TPN at a low target speed (25\%) in the straight and figure-8 target path, with the highest increase in hit rate as 0.28 among these configurations. This suggests that similar to the behavior of PN-Heading at low target speeds (further described in a below section), Hybrid TPN-Heading is able to chase the target at these speeds even if the initial TPN-driven approach is unsuccessful.

The LOS Guidance class of methods generally has higher hit rates than the Trajectory Following methods implemented here. When designing these algorithms and implementing them, it was found that significant parameter tuning and filtering was necessary to improve the results of the Trajectory Following methods. For example, although the LOS' Trajectory method produces smooth trajectories and therefore smoother UAV flight, it has to use a filtered (smoothed with flat moving average filter) LOS' in order to create consistent trajectories. This filtering introduces lag, which quickly becomes intractable when the target or UAV speed is increased past what was used for tuning the system. This is amplified in the case of the Forecasting Trajectory. While Equation \ref{eq:fortraj-p-collision} is geometrically sound, in the simulated system imperfections in depth estimation, LOS computation, and corresponding frame transformations (especially during high roll- and pitch-rates) cause inaccuracies that require filtering and tuning to make the 3D target forecast feasible. These limitations are largely a byproduct of using only monocular camera information to estimate 3D positioning. The effect can be seen in the UAV instability in the majority of experiments shown in Figure \ref{fig:fortraj-hit-rate}.

\subsubsection{TPN with Straight Target Path at Low Speeds}

\begin{figure}[H]
\centering
\subcaptionbox{\label{PN_slow_fail:a} Top-down view of UAV (RGB axes) approaching target (blue). Paths shown are approximately 5 seconds in length, with UAV path continuing off-screen.}{\includegraphics[width=.44\linewidth]{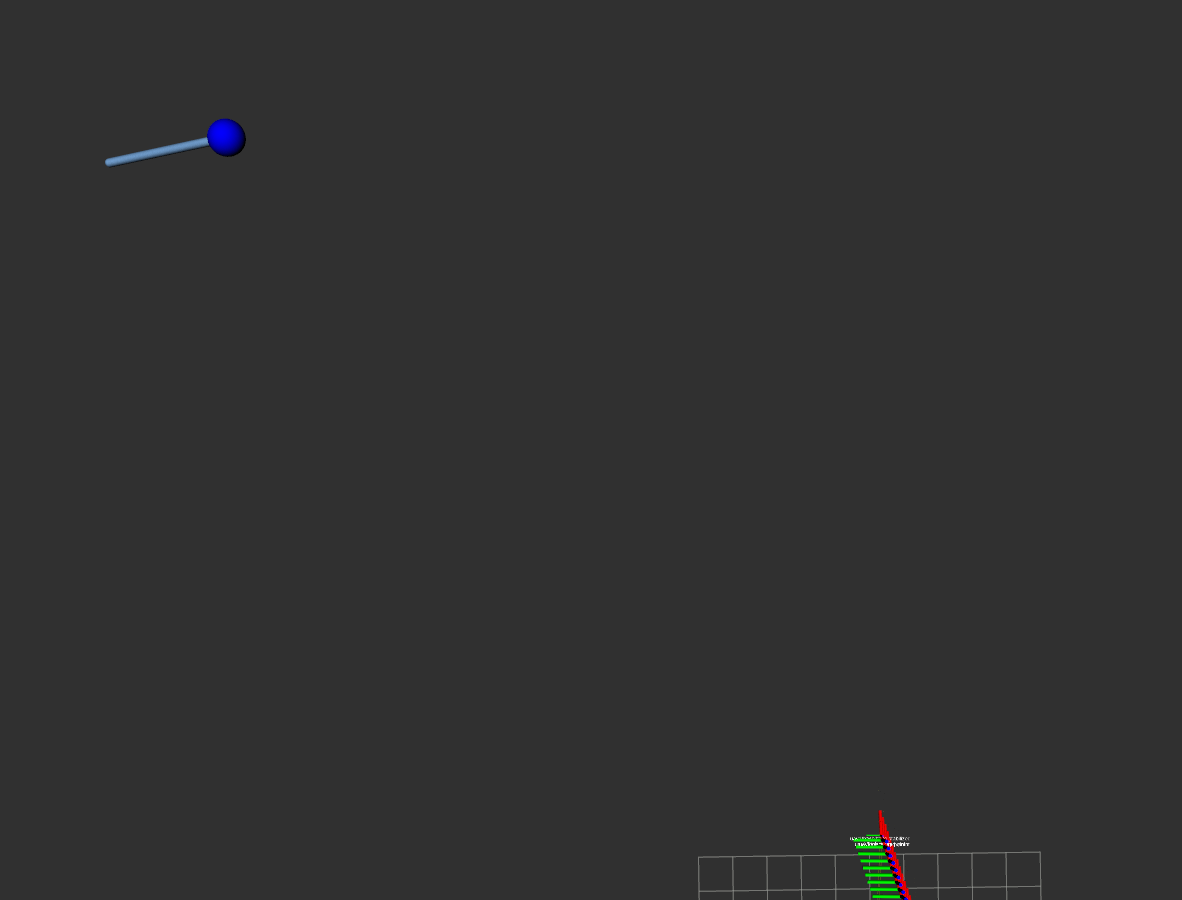}}\hfill
\subcaptionbox{\label{PN_slow_fail:b} Capture of image frame from UAV camera at the point shown in \ref{PN_slow_fail:a}.}{\includegraphics[width=.48\linewidth]{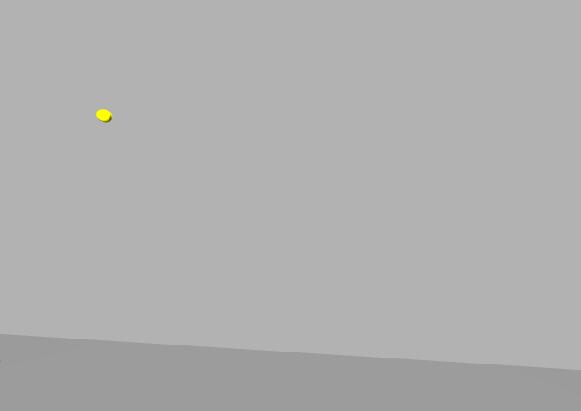}}\\
\subcaptionbox{\label{PN_slow_fail:c} Top-down view approximately 2s after \ref{PN_slow_fail:a}.}{\includegraphics[width=.44\linewidth]{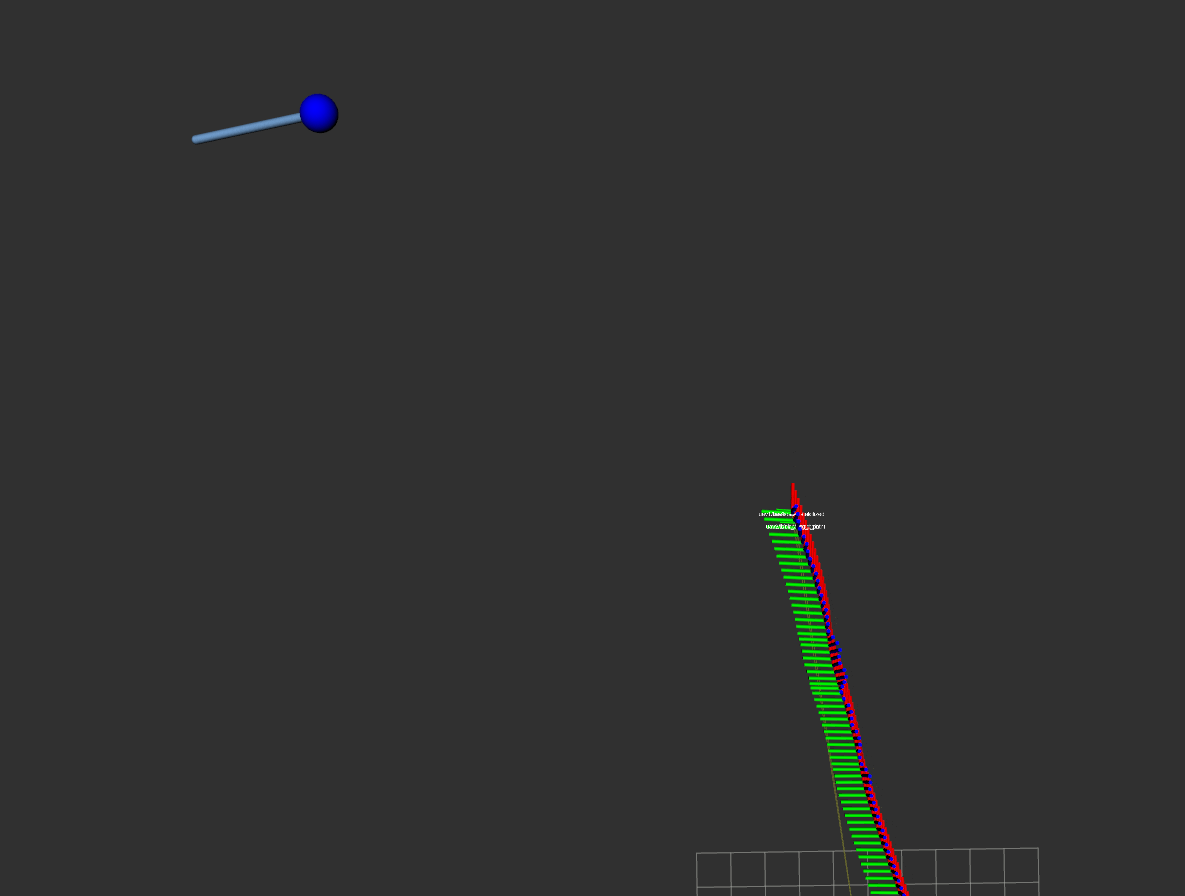}}\hfill
\subcaptionbox{\label{PN_slow_fail:d} Capture of image frame from UAV camera at the point shown in \ref{PN_slow_fail:c}.}{\includegraphics[width=.48\linewidth]{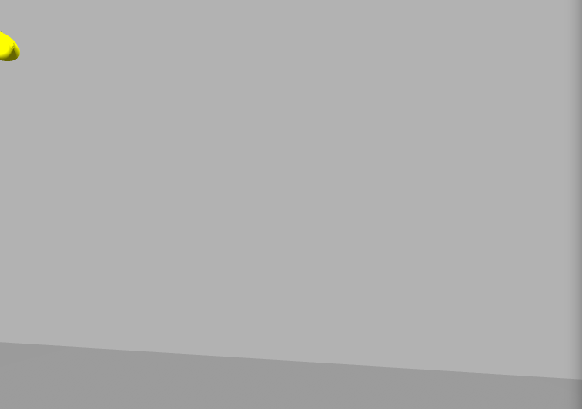}}\\
\subcaptionbox{\label{PN_slow_fail:e} Top-down view approximately 2s after \ref{PN_slow_fail:c}.}{\includegraphics[width=.44\linewidth]{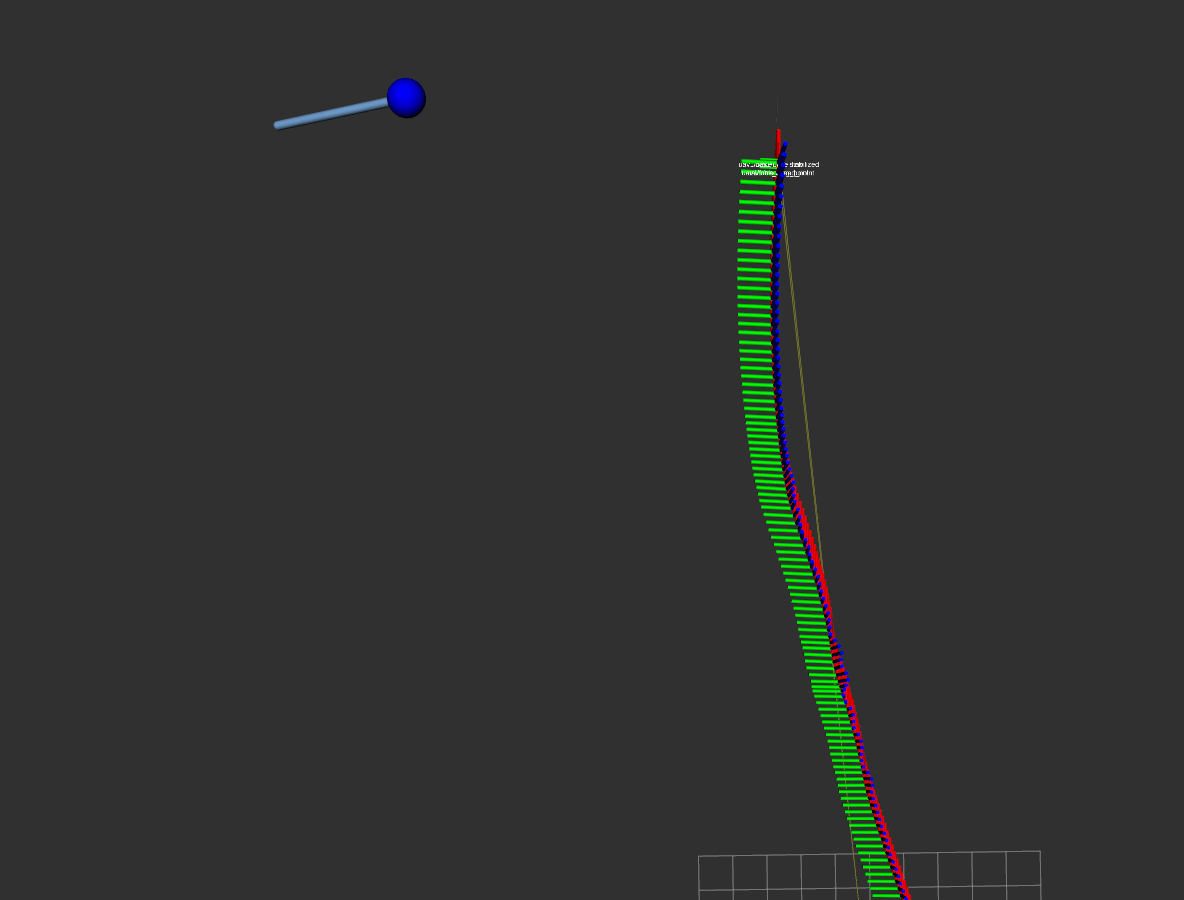}}\hfill
\subcaptionbox{\label{PN_slow_fail:f} Capture of image frame from UAV camera at the point shown in \ref{PN_slow_fail:e}.}{\includegraphics[width=.48\linewidth]{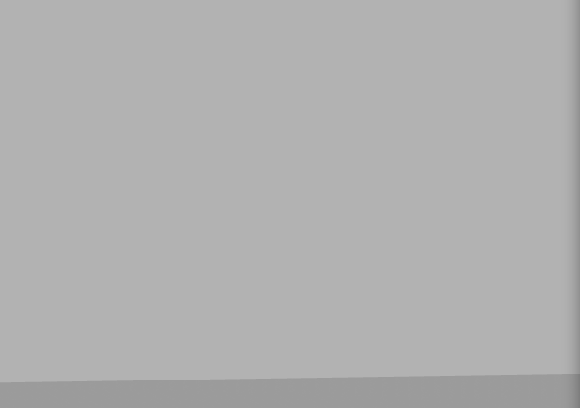}}\\
\caption{Demonstration of TPN, straight target path, UAV speed 2m/s and target speed 25\% (0.5m/s); corresponding datapoint is represented in top left corner of Figure \ref{tpn-hit-rate:a}.}
\label{fig:PN_slow_fail}
\end{figure}

While TPN generally achieves the highest hit rate compared to all other methods presented here, a notable exception is the slowest configuration of UAV velocity 2m/s and target speed 25\% with the straight target path, which has a hit rate of 0.48. Figure \ref{fig:PN_slow_fail} shows that the target progressively gets farther out of view as the UAV flies by. The straight target path begins on one side of the pre-defined arena space and terminates on the other side, with randomized variation in path slope. Since the target moves so slowly in this configuration, it remains on one side of the UAV's image. The $y$ component of the body-frame acceleration command generated by TPN while the target is moving towards the edge of the image (due to the UAV's forward velocity) is not enough to keep the target in sight, and it slowly slips out of view. This problem becomes less likely at higher UAV speeds, however, since the greater motion of the UAV creates more motion of the target in the image, thereby creating a larger lateral acceleration command that keeps the target in view. This issue is only apparent in the straight path case since only with this path the starting target location had to be on the edges of the image to compensate for larger movement at higher target speeds. This issue does not appear in the PN-Heading or Hybrid TPN-Heading methods, since they compensate by commanding yaw-rate to center the object in the image.

\subsubsection{PN-Heading at High Speeds}

PN-Heading has lower performance at high UAV or target speeds relative to both TPN or Hybrid TPN-Heading. This was often observed to be due to the time required to effectively change the UAV's velocity direction through a yaw-rate, and this time delay increases further as the UAV speed increases. A notable artifact in the results can be seen in Figure \ref{pnhc-hit-rate:a}, where the results at 100\% target speed is 0 at every UAV velocity. This is due to the way in which the UAV ``chases" the target in the straight target path scenario. Figure \ref{fig:pnhc-rviz} shows that the UAV might pass the target initially, but turns towards it via commanding a yawrate, and finds it again (within the 3s detection timeout). Then it is able to pursue the target with PN acceleration commands in the $x$ and $z$ directions. However, during this chase period, the UAV's velocity is in the same direction as the target's. Therefore, the target's speed must be less than the UAV's, or it will be impossible to maintain a nonzero closing velocity. This is the case specifically for the bottom row of Figure \ref{pnhc-hit-rate:a}. This does not occur with the other target motions since the target has non-zero acceleration.

% analyze PN-Heading at high UAV vel but low target vel

% maybe better results with fig8, knot at higher speeds bc the target reverses direction sometimes, as opposed to sl

% TODO explain and include figure of PN-Heading with 1.0 SL target

% "at high speeds multiple passes would likely work"

\subsection{Pursuit Duration}
\label{subsec:pursuit-duration}

\begin{figure}[H]
\centering
\subcaptionbox{\label{pd-sl:a}Pursuit duration vs. UAV velocity.}{\includegraphics[width=.47\linewidth]{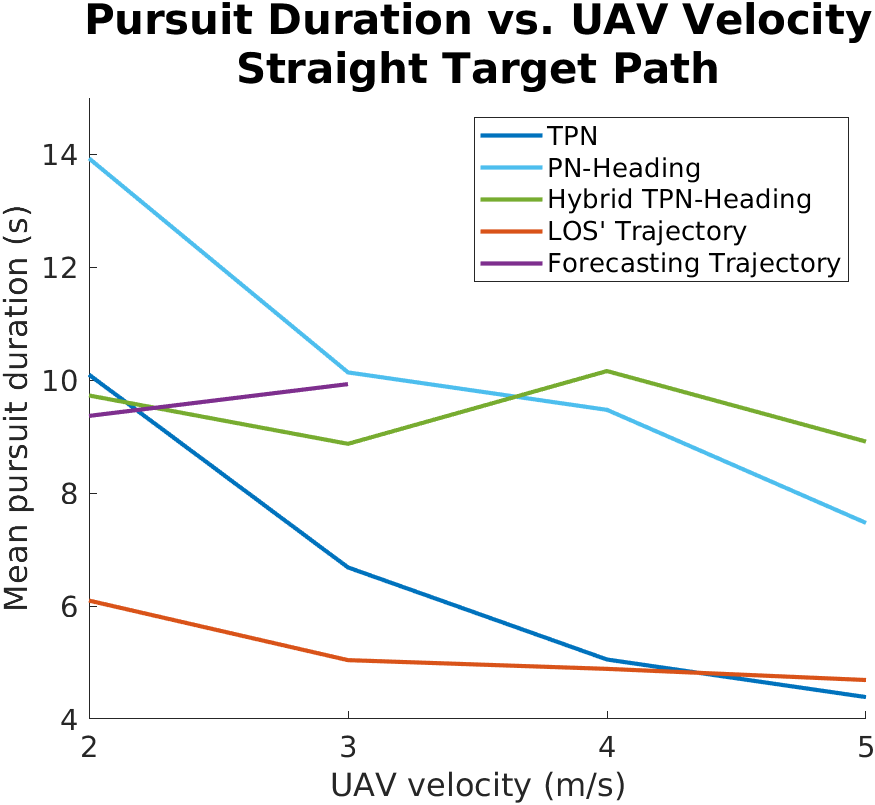}}\hfill
\subcaptionbox{\label{pd-sl:b}Pursuit duration vs. target speed.}{\includegraphics[width=.51\linewidth]{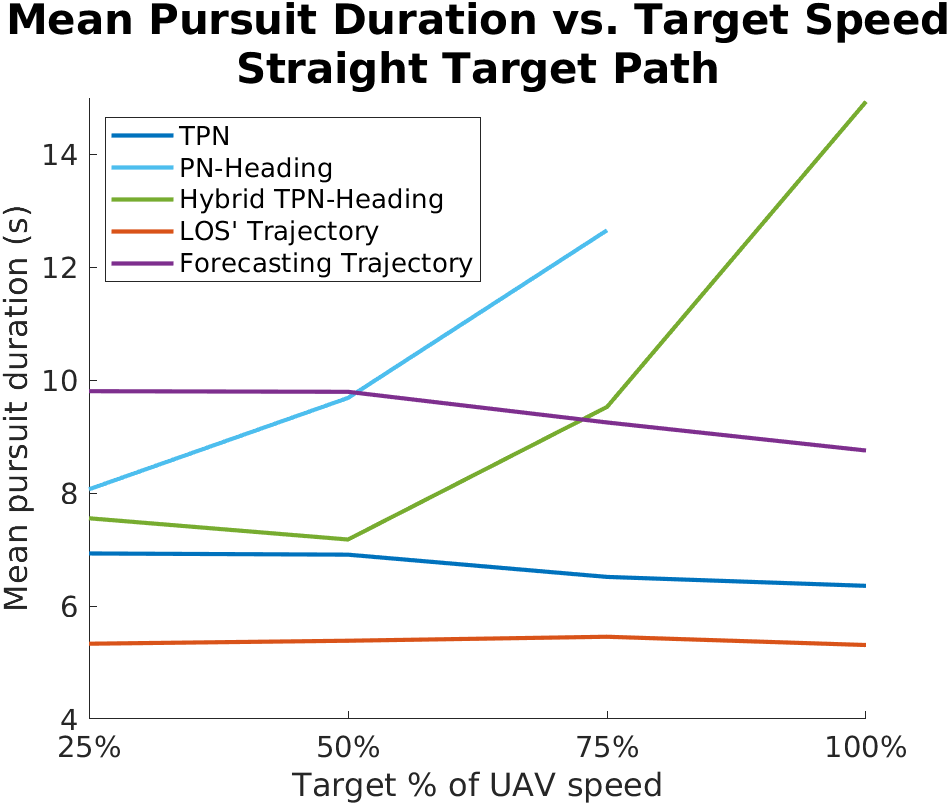}}\\
\caption{Mean pursuit durations for straight target path.}
\label{fig:pd-sl}
\end{figure}

If there were no hits across all experiments for a particular configuration, then those data points are missing in the figures, as seen in \ref{pd-sl:a} and \ref{pd-sl:b}.

\begin{figure}[H]
\centering
\subcaptionbox{\label{pd-fig8:a}Pursuit duration vs. UAV velocity.}{\includegraphics[width=.47\linewidth]{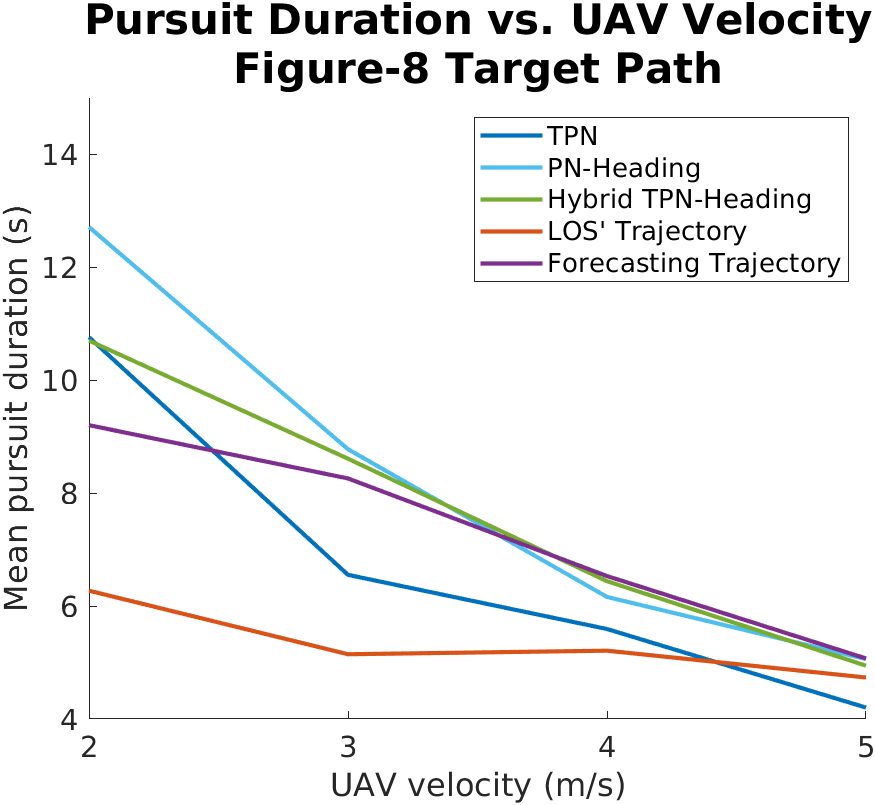}}\hfill
\subcaptionbox{\label{pd-fig8:b}Pursuit duration vs. target speed.}{\includegraphics[width=.51\linewidth]{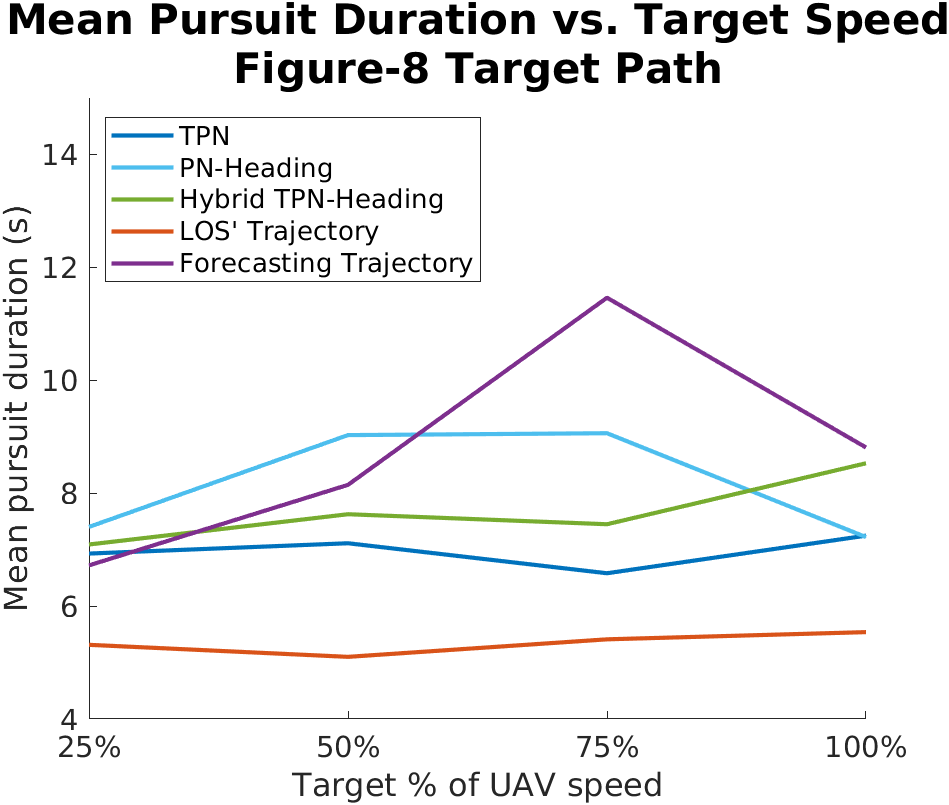}}\\
\caption{Mean pursuit durations for figure-8 target path.}
\label{fig:pd-fig8}
\end{figure}

\begin{figure}[H]
\centering
\subcaptionbox{\label{pd-knotl:a}Pursuit duration vs. UAV velocity.}{\includegraphics[width=.47\linewidth]{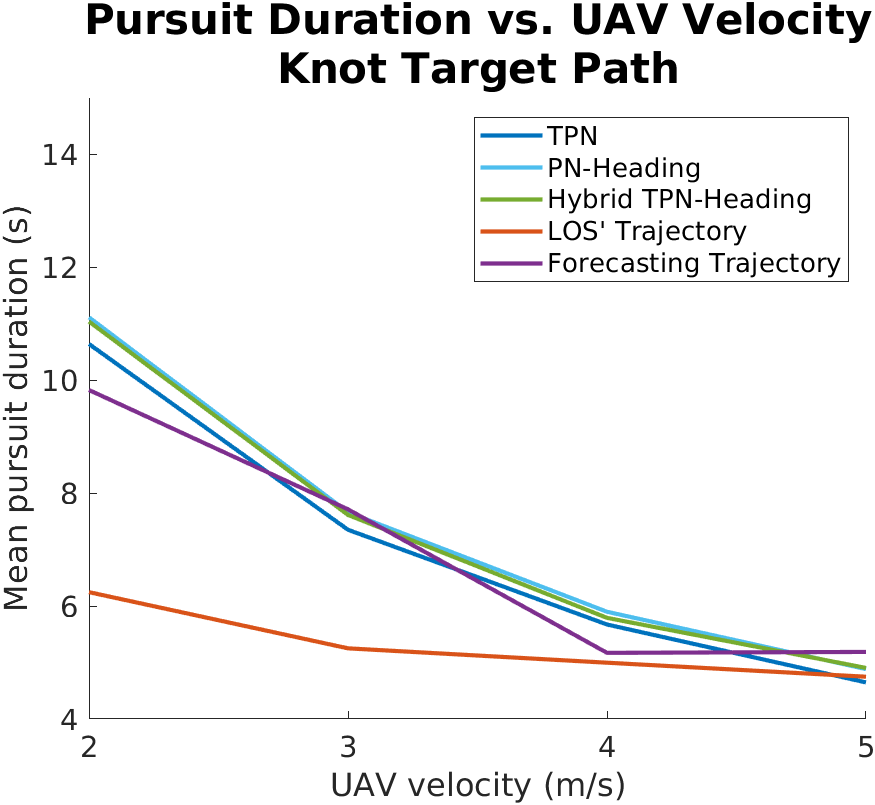}}\hfill
\subcaptionbox{\label{pd-knot:b}Pursuit duration vs. target speed.}{\includegraphics[width=.51\linewidth]{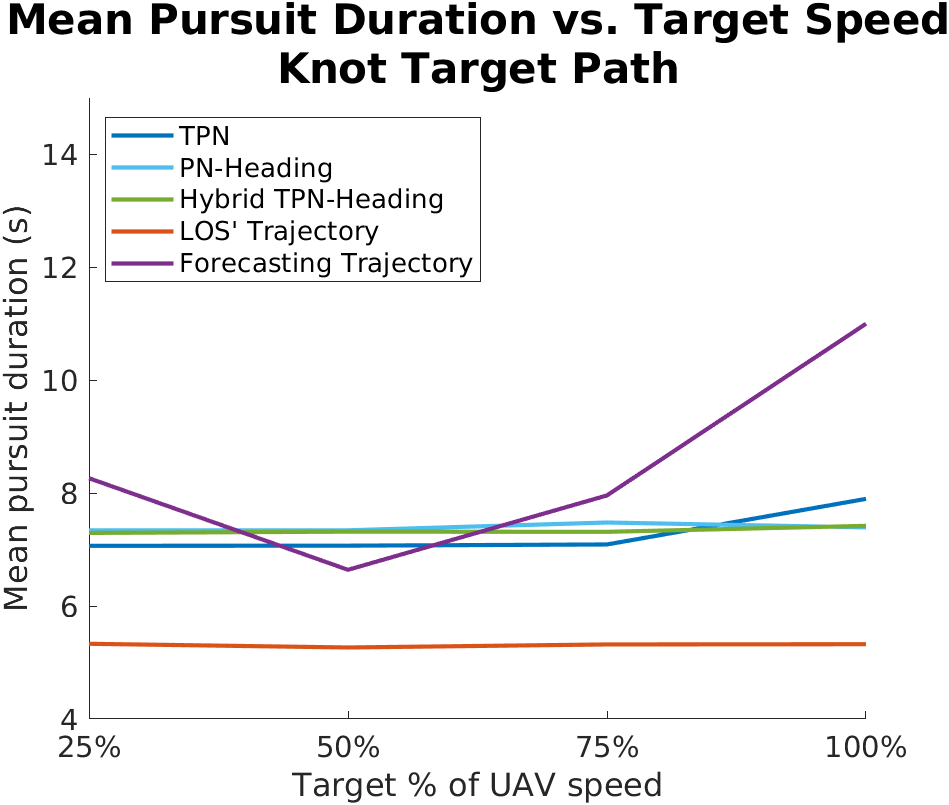}}\\
\caption{Mean pursuit durations for knot target path.}
\label{fig:pd-knot}
\end{figure}

The pursuit times generally decreased as UAV velocity increased, though this trend is not as apparent in some of the guidance methods in Figure \ref{pd-sl:a}. In the straight target path scenario, because the target has constant velocity it will eventually be out of the UAV's FOV, restricting the time window in which an intercept is possible. A strong downward trend is still apparent with TPN and PN-Heading.

The effect of target speed on pursuit duration is most observed in the straight target path case. Here, the increase in target speed caused increased times for PN-Heading and Hybrid TPN-Heading. Both of these methods utilize yaw-rate control to keep the target centered in the UAV's camera image, and as shown in Figure \ref{fig:pnhc-rviz}, this can result in the UAV turning and ``chasing" the target as it passes. As the target speed increases, the probability of a successful hit on first approach goes down, which then increases the chance of this kind of chasing maneuver.

LOS' Trajectory, though using a smoothed version of the LOS' used for TPN, does not exhibit as strong of a decrease in pursuit time over increasing UAV velocity or target speed. These results actually show that of all methods, LOS' Trajectory has the lowest pursuit durations with few exceptions. However, when paired with the results in Figure \ref{fig:lostraj-hit-rate}, it seems more likely that these times are a byproduct of being most likely to hit a target that is initialized closer to the UAV starting point.

\subsection{Pursuit Behavior}

In this section, we present each method's UAV path relative to target motion in hand-picked configurations for qualitative assessment.

% TODO generate these figures when computer doesn't randomly freeze on me

\begin{figure}[H]
    \centering
    \includegraphics[width=0.6\textwidth]{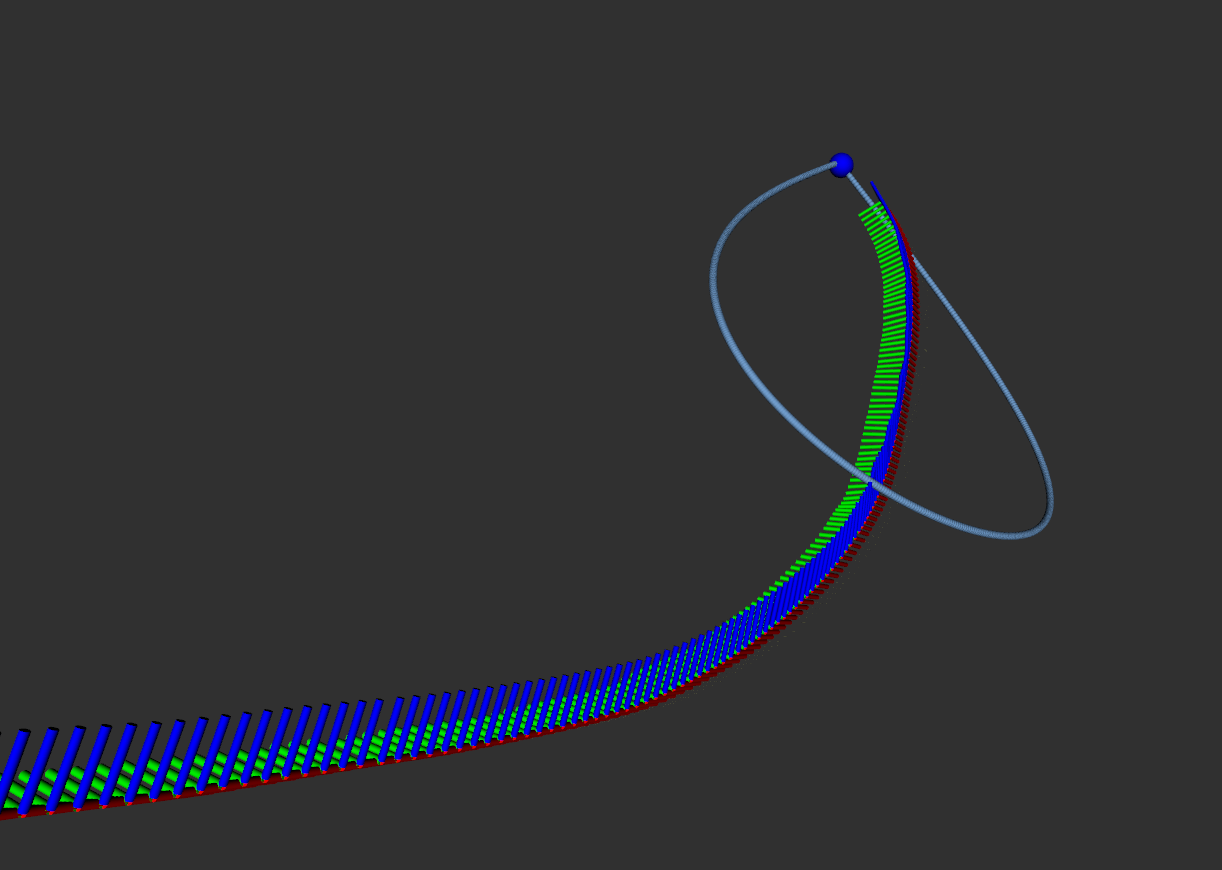}
    \caption{TPN behavior.}
    \label{fig:tpn-rviz}
\end{figure}

\begin{figure}[H]
    \centering
    \includegraphics[width=0.6\textwidth]{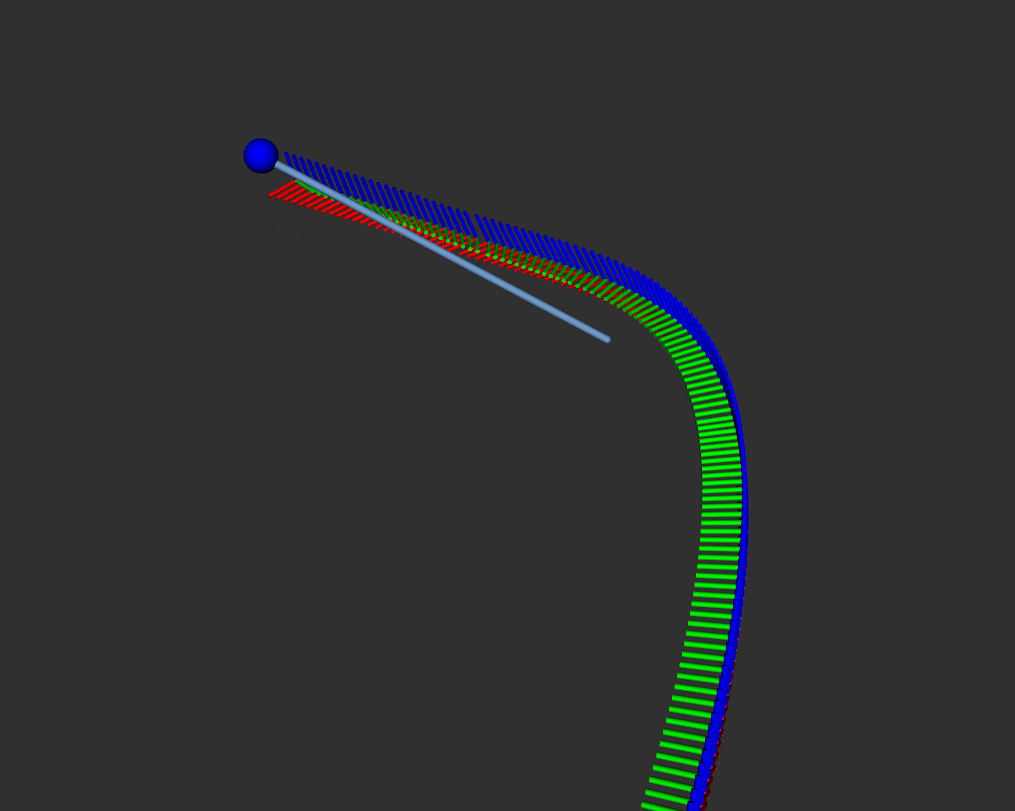}
    \caption{PN-Heading behavior.}
    \label{fig:pnhc-rviz}
\end{figure}

\begin{figure}[H]
    \centering
    \includegraphics[width=0.6\textwidth]{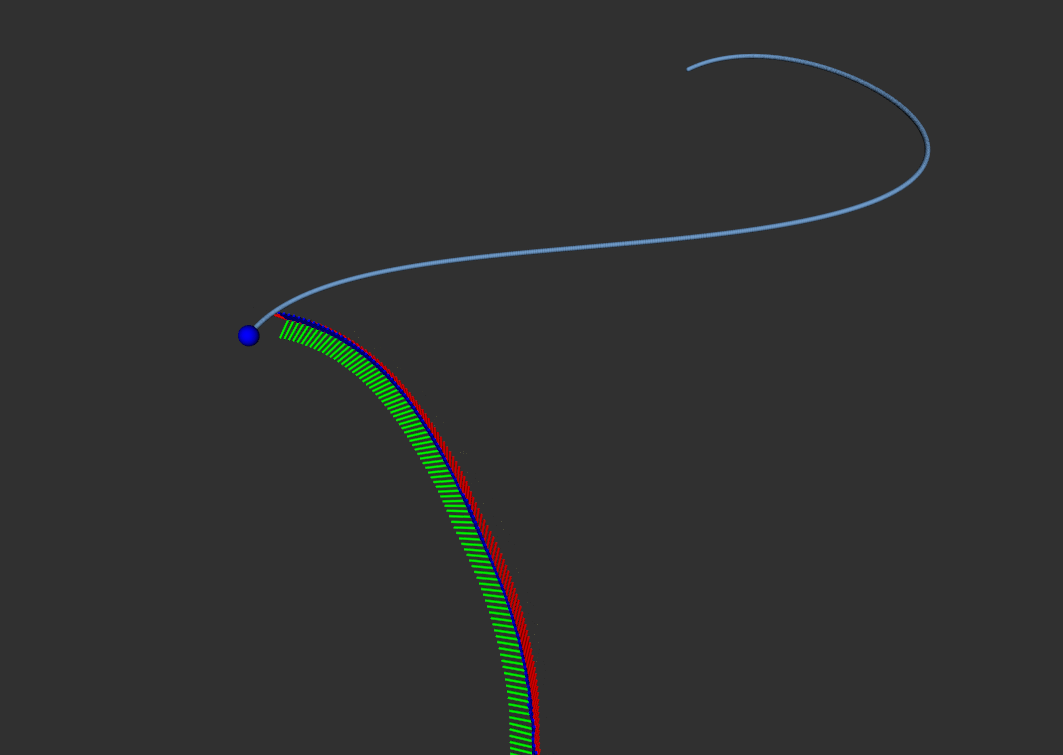}
    \caption{Hybrid TPN-Heading behavior.}
    \label{fig:hybrid-rviz}
\end{figure}

\begin{figure}[H]
    \centering
    \includegraphics[width=0.6\textwidth]{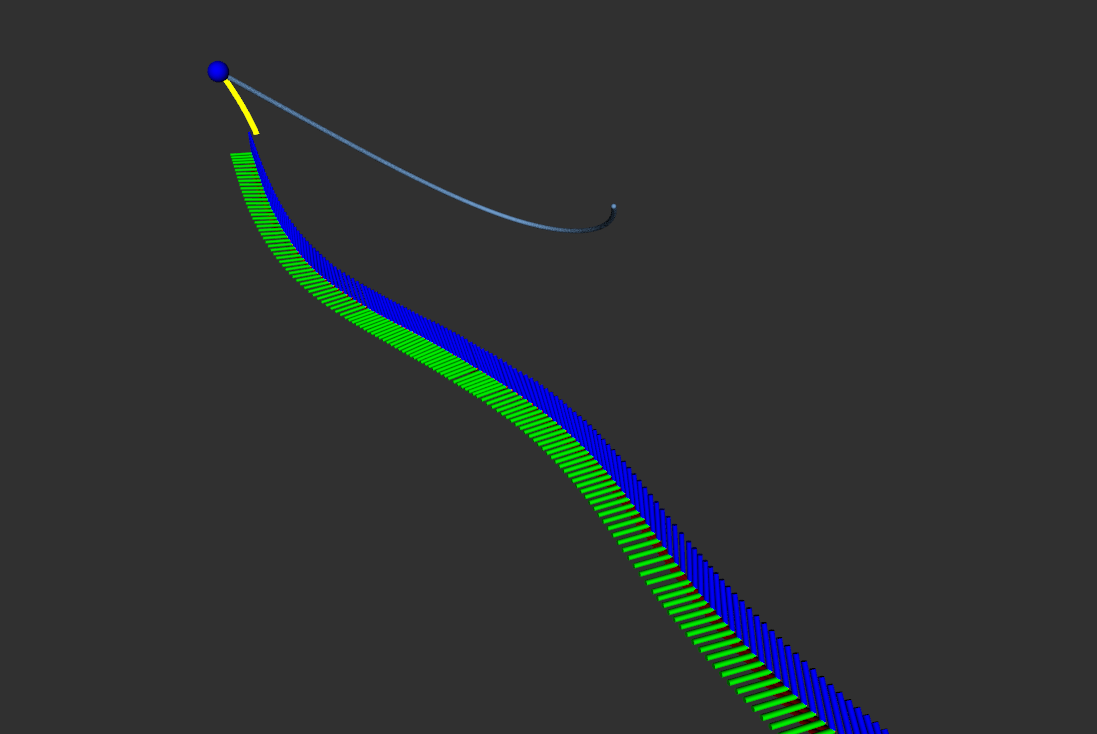}
    \caption{LOS' Trajectory behavior.}
    \label{fig:lostraj-rviz}
\end{figure}

\begin{figure}[H]
    \centering
    \includegraphics[width=0.6\textwidth]{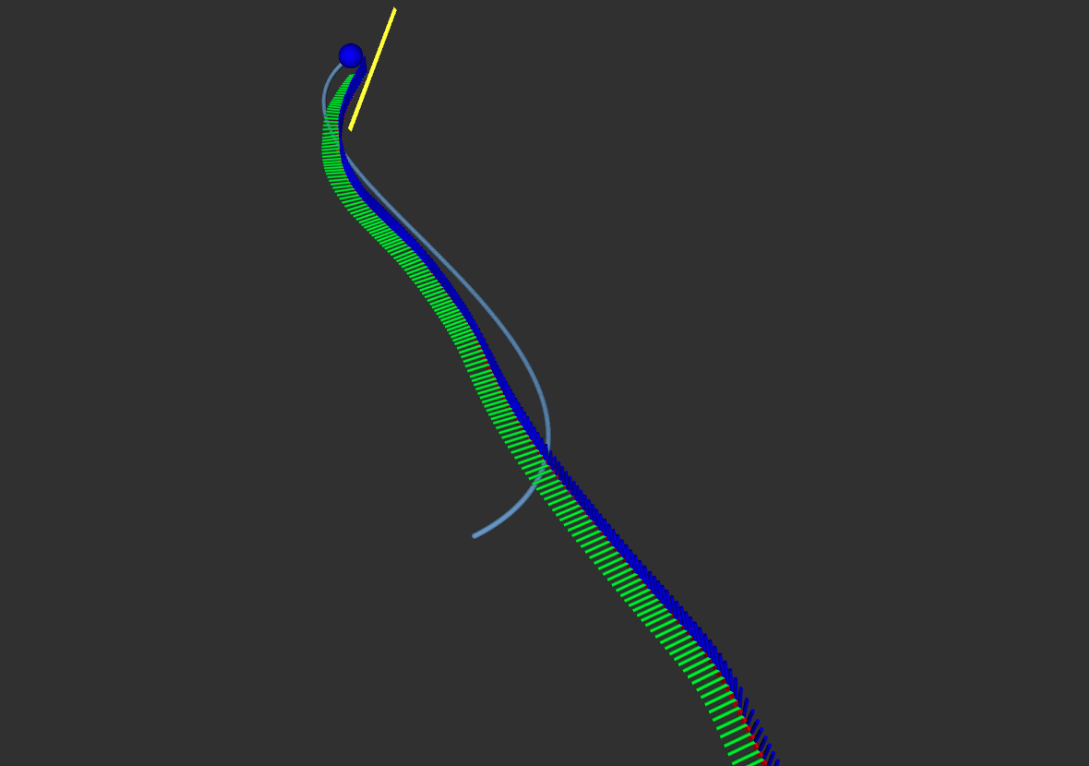}
    \caption{Forecasting Trajectory behavior.}
    \label{fig:fortraj-rviz}
\end{figure}

Figure \ref{fig:tpn-rviz} shows the UAV track the target's motion as it dips and then rises again as it moves through one half of the figure-8 path. Figure \ref{fig:pnhc-rviz} shows the UAV approach the straight target path but make a sharp left turn via yaw-rate commands as it passes by. It then goes on to implement PN to catch the target. Figure \ref{fig:hybrid-rviz} starts with motion similar to TPN, then uses heading control to yaw towards the target when it is close and the heading becomes larger. Figure \ref{fig:lostraj-rviz} shows a smoother path than TPN, as a result of smoothing the PN commands and utilizing trajectory following. The trajectory generated at the timestamp shown in the image is shown in yellow. Figure \ref{fig:fortraj-rviz} shows the path mimicking the motion of the target, but shifted in space due to forecasting of the target motion. The current forecasted trajectory is shown in yellow.

 \chapter{LOS Guidance Applied in a Robotics Competition Setting}
\label{chap:mbzirc}

\section{Introduction}
\label{sec:mbz-intro}

The Mohamed Bin Zayed International Robotics Challenge 2020 is an outdoor robotics competition in which dozens of international teams, including many top robotics universities, demonstrate autonomous performance in different tasks.

\begin{quote}
    ``MBZIRC aims to provide an ambitious, science-based, and technologically demanding set of challenges in robotics, open to a large number of international teams. It is intended to demonstrate the current state of the art in robotics in terms of scientific and technological accomplishments, and to inspire the future of robotics." \cite{mbzirc}
\end{quote}

Teams had the choice of competing in any or all of three challenges, differentiated by the types and number of robots allowed, the theme of the tasks involved, and physically separated arenas. Challenge 1: Airspace Safety involved aerial robots targeting semi-stationary and moving targets; Challenge 2: Construction involved both ground and aerial robots building separate wall structures with provided blocks of varying size and weight; Challenge 3: Firefighting involved ground and aerial robots cooperating to extinguish real and fake fires surrounding and inside a three-story building. The work towards competing in Challenge 1 and the associated competition results are the most relevant towards the thesis of airspace safety, and are therefore the focus in this report.

In Challenge 1, the two primary tasks are as follows:

\begin{itemize}
    \item[] \textbf{Task 1: Semi-stationary targets.} Pop five 60cm-diameter green balloons placed randomly throughout the arena.
    \item[] \textbf{Task 2: Moving target.} Capture and return a 15cm yellow foam ball hanging from a UAV flying throughout the arena in a figure-8 path.
\end{itemize}

The balloons were placed on rigid poles but filled with helium and attached to the pole with string; therefore, they can move in a limited range due to wind or downdraft from a UAV's propellers. The moving target moved at speeds up to 8m/s along a figure-8 trajectory at some unknown altitude. After 8 minutes, the speed of the target was reduced to 3m/s. The ultimate goal is to remove the ball from the magnetic tether and return it to a dropoff location in the arena. The total time allowed for this challenge was 15 minutes, without the possibility of pausing the clock during robot resets. There was a choice of running the robots in autonomous or manual mode (scores in either category are compared separately), but all competition runs by Team Tartans were completed in autonomous mode. Each team was allowed to have at most three aerial robots in action at any given time in the arena. We used two UAVs, one for each of the two tasks listed above. The discussion in this thesis focuses on the most developed work done in the efforts towards this competition; as such, the entire software stack is discussed for Task 1, but only the target capture (not return) portion of Task 2 is covered.

% TODO example diagram of Challenge 1 arena with 5 balloons and flying drone+ball

\section{Software System}
\label{sec:mbz-system}

\begin{figure}[h!]
    \centering
    \includegraphics[width=0.7\textwidth]{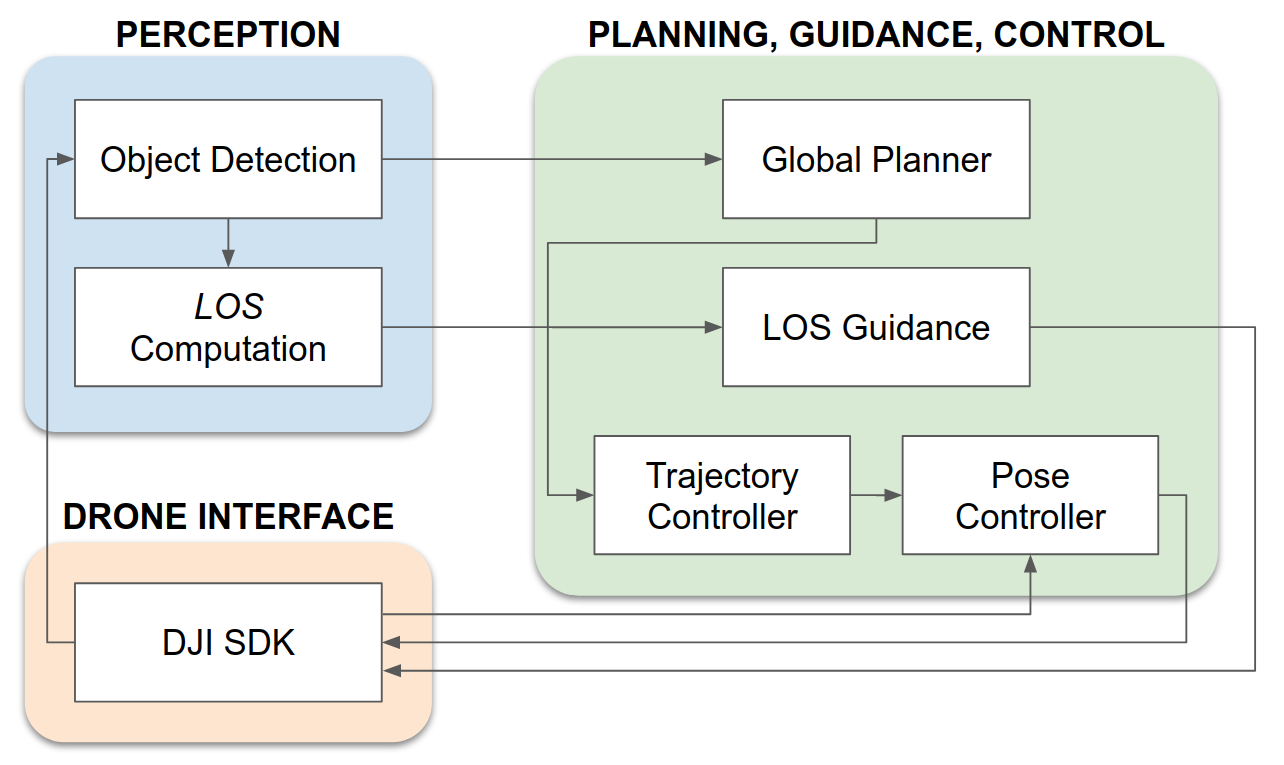}
    \caption{Diagram of software system used on both Challenge 1 UAVs.}
    \label{fig:mbz-software-system}
\end{figure}

Both UAVs used for Tasks 1 and 2, respectively, were very similar in their software stacks for perception, planning/guidance/control, and interface with the DJI SDK. The system can be simplified to the diagram seen in Figure \ref{fig:mbz-software-system}.

\subsection{Perception}
\label{subsec:mbz-system-perception}

\subsubsection{Object Detection}

The target objects in both tasks (balloons and ball) have distinct color and shape. For this reason, classical object detection and segmentation techniques such as color thresholding and ellipse fitting were tried due to their low computation and ease of implementation. However, through multiple trials in different environments (sunny, cloudy, grassy or urban backgrounds) the performance of these detectors failed unless hand-tuned for each case. We therefore turned to a deep learning approach, and evaluated different object detection architectures for accuracy and speed. The result was use of the Tiny YOLOv3 architecture, which provided high accuracy and real-time inference when paired with the Intel OpenVino framework. This framework optimizes the architecture to run on a CPU.

The accuracy of the network inference was increased by tuning hyperparameters during training, as well as augmenting the hand-collected dataset. Data of sample balloons and balls were collected in different settings (indoor/outdoor, sunny/cloudy, with/without obstruction or multiple objects in scene). Data augmentation also helped to increase the size and variation in the dataset.

% TODO sample training images | sample augmented training images

% TODO camera image of multiple balloons in sight with multiple detections but one target

Since there are multiple balloon targets placed throughout the arena, it is possible that in any given camera frame there may be multiple registered, true positive detections. To eliminate ambiguity in the balloon detection and targeting pipeline, certain conditions must be met for a detection to be considered a valid target. The primary condition is that the total area of the bounding box must be greater than some threshold. This parameter was specified as a percentage of the total image area, and was tuned during competition according to the size of the balloons used.

% TODO picture of ball target in frame with balloons below

Though two networks were trained and used for each of the two tasks, and each task's objects were colored differently (green balloons vs. yellow ball), false positives would often result from detections of the other task's targets. The most common case was the ball-trained network detecting balloons, since the Task 2 UAV had a much wider view of the arena from the higher altitude than the Task 1 UAV, which stayed under a 5m artificial ceiling. This problem was mitigated during the competition by filtering out detections from the ball-trained network based on location in the image; detections in the bottom 30\% of the image were discarded since the ball was expected to be at or above the Task 2 UAV's altitude.

\subsubsection{$LOS$ Computation}

This computation was performed with camera parameters obtained from the calibrated gimbal camera, published by the DJI SDK. Please refer to Equation \ref{eq:los} for the calculation. The LOS vector was then passed to the LOS Guidance node.

\subsection{Planning, Guidance, and Control}
\label{subsec:mbz-system-guidance}

\subsubsection{Global Planner}

\begin{figure}[h!]
    \begin{minipage}{.5\textwidth}
    \centering
    \includegraphics[height=4cm]{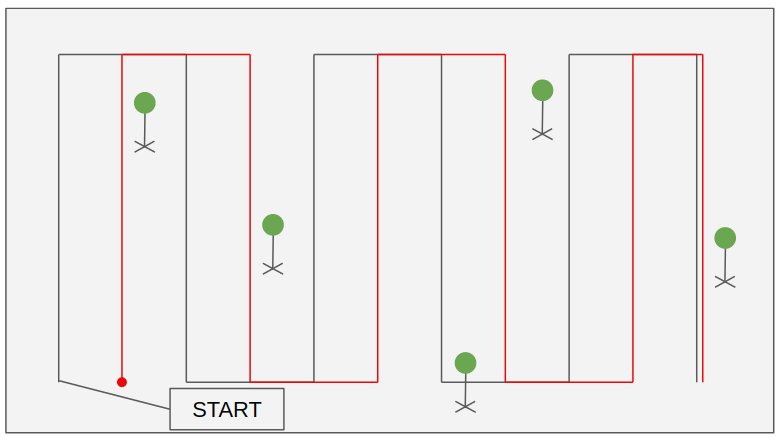}
    \caption{Global planner trajectory generated to search for balloons over the entire arena for Task 1. Black lines indicate a forward pass starting from the START position, and red lines indicate the shifted reverse pass.}
    \label{fig:drawn-global-plan-task1}
    \end{minipage}%
    \centering
    \begin{minipage}{.5\textwidth}
    \centering
    \includegraphics[height=4cm]{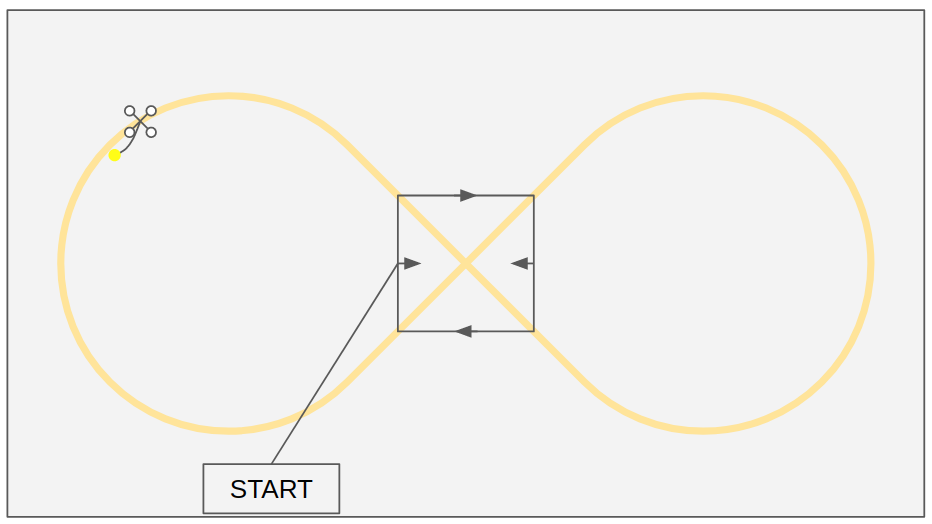}
    \caption{Global planner trajectory generated to search for the ball in the center of the arena for Task 2. Black lines indicate the pursuer UAV trajectory, with the black arrows indicating yaw orientation (and therefore camera orientation). The yellow line is the target UAV and ball path.}
    \label{fig:drawn-global-plan-task2}
    \end{minipage}
\end{figure}

The Global Planner produced a global trajectory search pattern for both Tasks 1 and 2. For Task 1, this consisted of a forward-backward lawnmower path running over a specified portion of the arena, at a preset altitude. It stitches separate trajectory segments together to create a complete trajectory including takeoff, recovery after LOS Guidance, and landing. The sweep width, altitude, and velocity were tuned during competition (e.g. 6m sweep width, 2.4m altitude, and 2m/s velocity). After LOS Guidance is triggered due to a valid target detection, the Global Planner pauses the trajectory. Once the state machine returns to the Global Planner, a segment is stitched to the original search pattern starting from the current UAV odometry, going up 1.5m, then going back to the original point at which the original detection was triggered. This is a simple way to ensure robustness in the balloon-popping pipeline.

For Task 2, the Global Planner produced a trajectory consisting of a square path in the center of the arena at a higher altitude than the Task 1 UAV. The orientation of the UAV was set to always align with the long side of the arena, facing towards the cross portion of the figure-8 path of the target. This ensured that the UAV should have the target both in view and in close proximity for the longest time possible. Once the target was identified, the state machine passes into LOS Guidance mode, and the UAV only returns to the Global Planner mode if the target has not been seen for the approximate time it takes for the target to complete two passes. The altitude of the search pattern was set to 10-12m during competition, with a slower velocity of 1m/s compared to Task 1. The wait time before transitioning from LOS Guidance mode to Global Planner was 60s, sufficient time for two passes of the target.

\subsubsection{$LOS$ Guidance}

Both Tasks 1 and 2 use the LOS vector to guide the UAV towards targets. For Task 1, once a valid detection is registered, as specified in Section \ref{subsec:mbz-system-perception}, the UAV enters LOS Guidance mode which attempts to puncture the target balloon with the propellers. This guidance mode runs through two states: Adjust and Attack. In the first state, Adjust, the node outputs z-velocity (altitude) and yawrate commands to satisfy two conditions: (i) the upwards angle of the LOS vector (computed in the camera frame) is within 5 degrees of a set angle threshold, and (ii) the horizontal angle of the LOS vector is within 5 degrees of 0. Once these conditions are met, the guidance transitions to the Attack state, where the UAV satisfies velocity commands along the LOS until a certain time period after the target has lost sight. This relies on the assumption that no targets will be placed sufficiently close to any obstacles or boundaries to require obstacle avoidance immediately after popping the balloon. The Adjust state satisfies these conditions so that when the UAV moves along the LOS, the upwards trajectory makes it more likely that the propellers will hit the balloon without causing a miss due to propeller downdraft. After the Attack state is complete, control is passed back to the Global Planner.

In Task 2, once a valid detection is registered, LOS Guidance issues velocity proportional to the LOS vector in the y- (lateral) and z-direction (altitude). The x-direction (forward) component of the velocity vector is 0. This adjusts the UAV position to align with the target's path. Once the UAV loses sight of the target, it transitions to the Wait state and waits in place for the next pass. If the target is not seen within some period of time (tuned during competition to be approximately the duration of two cycles of the target's path) then it will pass control back to the Global Planner. Otherwise, if the target is seen again, then velocity commands are again issued along the LOS to further refine the UAV position.

\subsubsection{Trajectory and Pose Controllers}

These controllers are the same as used for the previous portion of this thesis, so please refer to Chapter \ref{chap:moving_objects} for additional details. Since the DJI SDK node contains an internal velocity controller and therefore accepts velocity commands, only Trajectory and Pose Controllers were used for this platform.

\subsection{Drone Interface}
\label{subsec:mbz-system-dji-sdk}

\subsubsection{DJI SDK}

The DJI Onboard SDK was used for interfacing with the UAV sensors and autopilot. Specifically, the $\texttt{dji\_sdk}$ ROS package was used with the specific SDK version corresponding to the DJI M210 V2. It should be noted that while the SDK supported gimbal camera angle control, a custom gimbal angle controller was never successful in controlling the gimbal angle when the UAV was yawed 180 degrees from the initial UAV orientation. Because of this, a custom gimbal angle controller was never able to be used, and the team relied on the internal SDK controller to maintain the preset angle set relative to the UAV. This sometimes failed mid-flight, which on at least one occasion specifically caused missed targets in Task 1. It was later found that there was a known bug in the SDK's handling of gimbal angle commands and angle wrapping past 180 degrees.

\section{Results}
\label{sec:mbz-results}

\subsection{Task 1}
\label{subsec:mbz-results-task1}

The competition trials were split over the course of three days, with the first two days focused on individual challenge trials and the third day hosting the grand challenge, when all three challenges are run at once under modified rules. For the purpose of this thesis, we will consider every trial on any of these days equally.

\begin{table}[h!]
\centering
\caption{Success tally for Task 1 subtasks across all competition trials.}
\label{tab:task1-subtasks}
\begin{tabular}{|c|c|c|}
\hline
\textbf{Subtask}      & \textbf{Success count} & \textbf{Success rate} \\ \hline
Target identification & 24/26                  & 92\%                  \\ \hline
Pop sequence          & 15/25                  & 60\%                  \\ \hline
Recover \& pop        & 4/4                    & 100\%                 \\ \hline
\end{tabular}
\end{table}

Table \ref{tab:task1-subtasks} shows the complete tally for three subtasks for Task 1 over all competition trials. Note that this excludes only one trial run from competition, which is excluded because of bad gimbal angle initialization during system setup (refer to Section \ref{subsec:mbz-system-dji-sdk}). Table \ref{tab:task1-failures} specifies the failure cause for each of the Task 1 subtask failures.

The \textit{Target identification} subtask was defined by the UAV reaching the closest point on the global search plan to a particular balloon and registering it as a valid target. There were two failures in this subtask. In both cases, though the incidents are from different trials, the reasoning was the same: the balloons happened to be outside the search path bounds and therefore appeared too small in the image to pass the required area threshold described in Section \ref{subsec:mbz-system-perception}. This was largely a failure in correctly setting global search path.

% TODO camera frame with target too far out of bounds

The \textit{Pop sequence} subtask was defined as, after having registered a target detection, successfully adjusting and guiding the UAV such that the balloon is popped. The overwhelming cause for a low success rate for this subtask was the detection of a balloon during a turn in the search path. Turns were tuned to be executed quickly to ensure that the 40m$\times$100m arena is covered efficiently. However, registering a detection during these turns almost always resulted in a failed pop sequence, since the internal DJI SDK delay in camera feed of ~0.1s resulted in a delayed notification of a target after the UAV had turned too far to recover. Other failure cases for the pop sequence included the gimbal being off-center and thereby creating an inaccurate LOS vector, and instances of propeller downdraft moving the balloon away from the propellers.

% TODO camera frames 1, 2, 3, in sequence showing a high-rate turn with labels on frames

\begin{table}[h!]
\centering
\caption{Causes of failure for the Task 1 subtasks.}
\label{tab:task1-failures}
\begin{tabular}{|c|c|c|}
\hline
\textbf{Subtask}       & \textbf{Cause of failure}      & \textbf{Count} \\ \hline
Target identification  & Search pattern far from target & 2              \\ \hline
\multicolumn{1}{|l|}{} & High-rate turn when detected   & 6              \\ \cline{2-3} 
Pop sequence           & Gimbal off-center              & 2              \\ \cline{2-3} 
\multicolumn{1}{|l|}{} & Propeller downdraft            & 2              \\ \hline
Recover \& pop       & N/A                            & N/A            \\ \hline
\end{tabular}
\end{table}

% \begin{table}[h!]
% \centering
% \caption{Causes of failure for the \textit{Pop sequence} subtask.}
% \label{tab:pop-sequence-failures}
% \begin{tabular}{|c|c|}
% \hline
% \textbf{Cause of failure}    & \textbf{Count} \\ \hline
% High-rate turn when detected & 6              \\ \hline
% Gimbal off-center            & 2              \\ \hline
% Propeller downdraft          & 2              \\ \hline
% \end{tabular}
% \end{table}

\begin{figure*}[ht!]
   \subfloat[\label{fig:balloon-detect}Target identification.]{%
      \includegraphics[trim=20 40 50 30,clip, height=3.3cm]{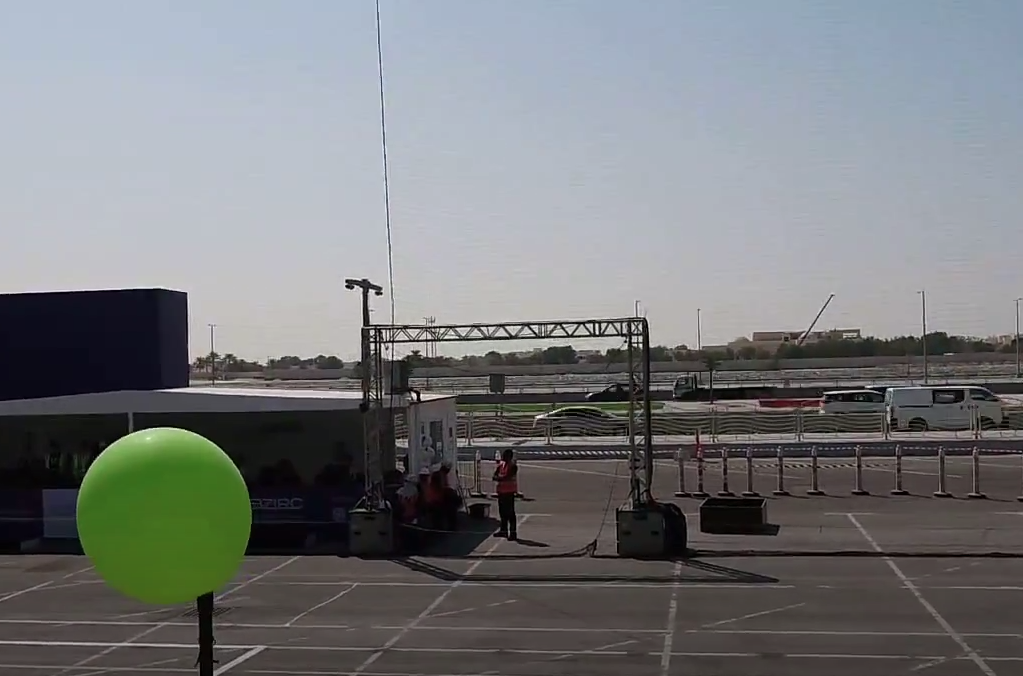}}
\hspace{\fill}
   \subfloat[\label{fig:balloon-adjust}Completion of Adjust portion of Pop sequence.]{%
      \includegraphics[trim=10 40 10 30,clip, height=3.3cm]{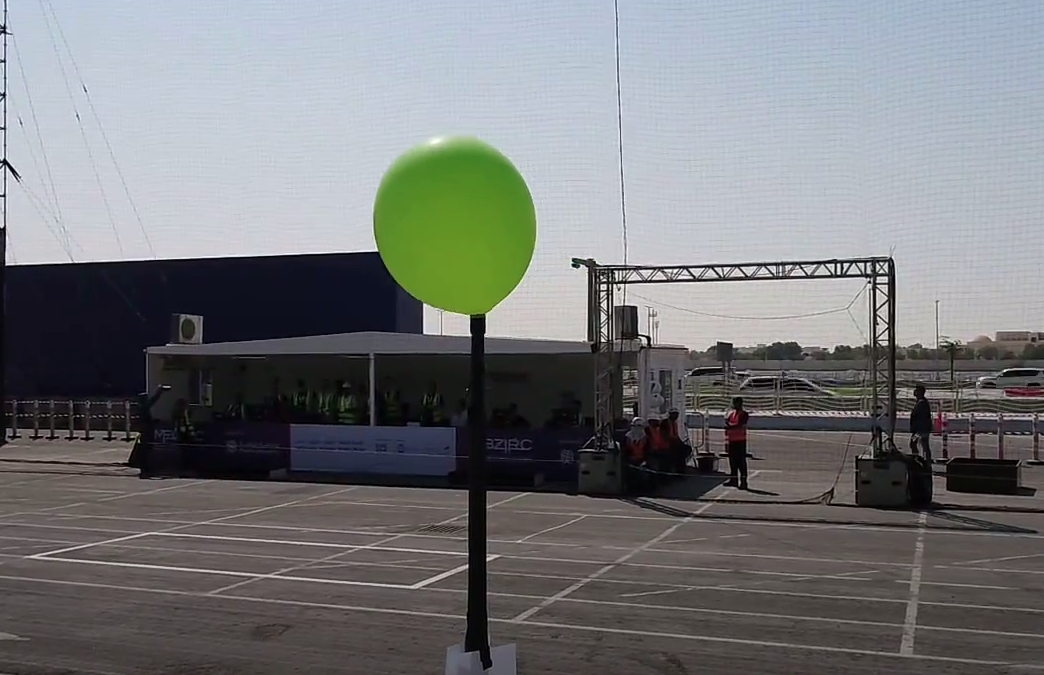}}
\hspace{\fill}
   \subfloat[\label{fig:balloon-recover}Recovery to original position.]{%
      \includegraphics[trim=30 40 30 30,clip, height=3.3cm]{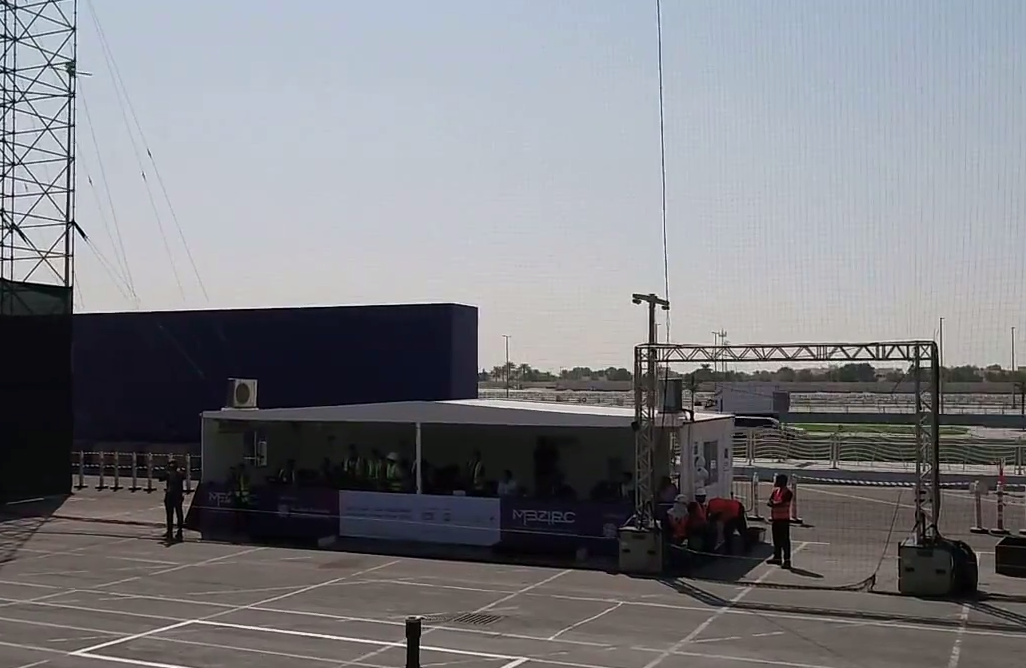}}\\
\caption{\label{fig:task1-snapshots}Snapshots of Task 1 subtasks.}
\end{figure*}

The \textit{Recover \& pop} subtask was defined as the pop sequence attempt after having failed the first pop attempt on a target and resuming the initial global search position from which the target was seen. In the case of the pop failures resulting from high-rate turns, the LOS Guidance state machine never transitioned into Attack (refer to Section \ref{subsec:mbz-system-guidance}); therefore, there was no recovery and second attempt in these cases. However, the four failures caused by either an off-center gimbal camera and downdraft from the propellers did create an opportunity to recover, view the target again, and try again. In all four recovery and pop attempts, the UAV was successful in popping the balloon.

\subsection{Task 2}
\label{subsec:mbz-results-task2}

Due to complications during the competition, there are only two trials from which to report results for Task 2. As such, the analysis of Task 2 performance is confined to qualitative results. During both of these trials, the UAV was able to find the yellow ball target via the square search pattern shown in Figure \ref{fig:drawn-global-plan-task2}. However, since the direction in which the target is flying the figure-8 path is not previously known, this particular search plan can spot the target either when it is approaching or receding from the UAV. In both trials, the UAV successfully identified the target and implemented LOS guidance towards it, but did so when it was flying away from the UAV. This prevented the UAV from ever achieving the position necessary to intercept the target, as described in geometric detail below.

\begin{figure*}[ht!]
   \subfloat[\label{fig:target-frame1}]{%
      \includegraphics[trim=20 40 50 30,clip, width=0.34\textwidth]{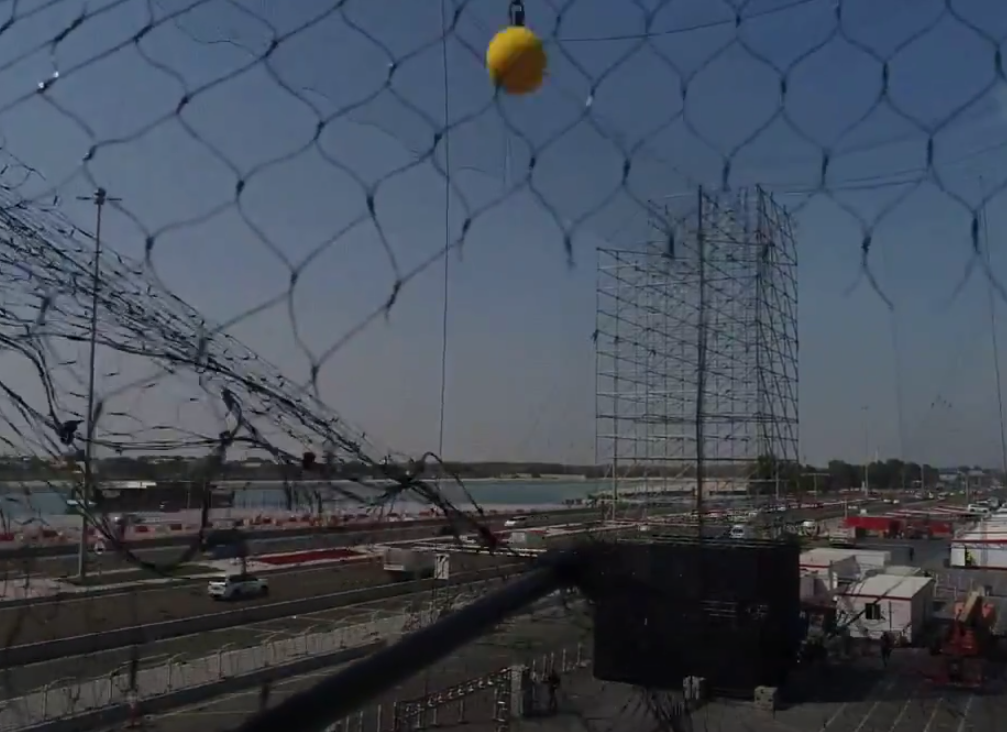}}
\hspace{\fill}
   \subfloat[\label{fig:target-frame2} ]{%
      \includegraphics[trim=10 40 10 30,clip, width=0.33\textwidth]{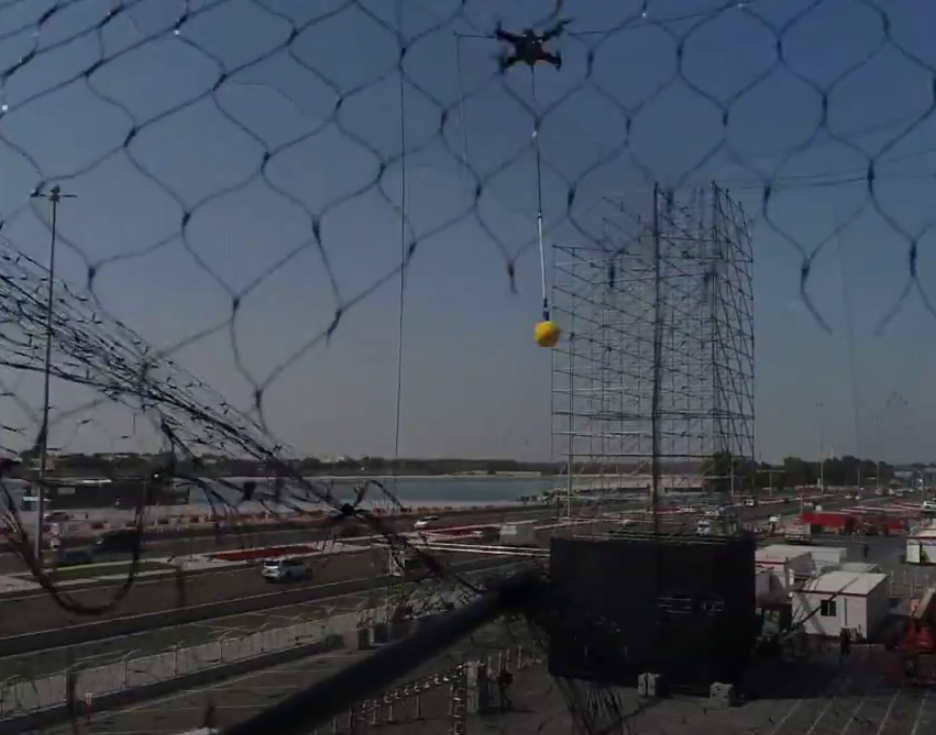}}
\hspace{\fill}
   \subfloat[\label{fig:target-frame3}]{%
      \includegraphics[trim=30 40 30 30,clip, width=0.32\textwidth]{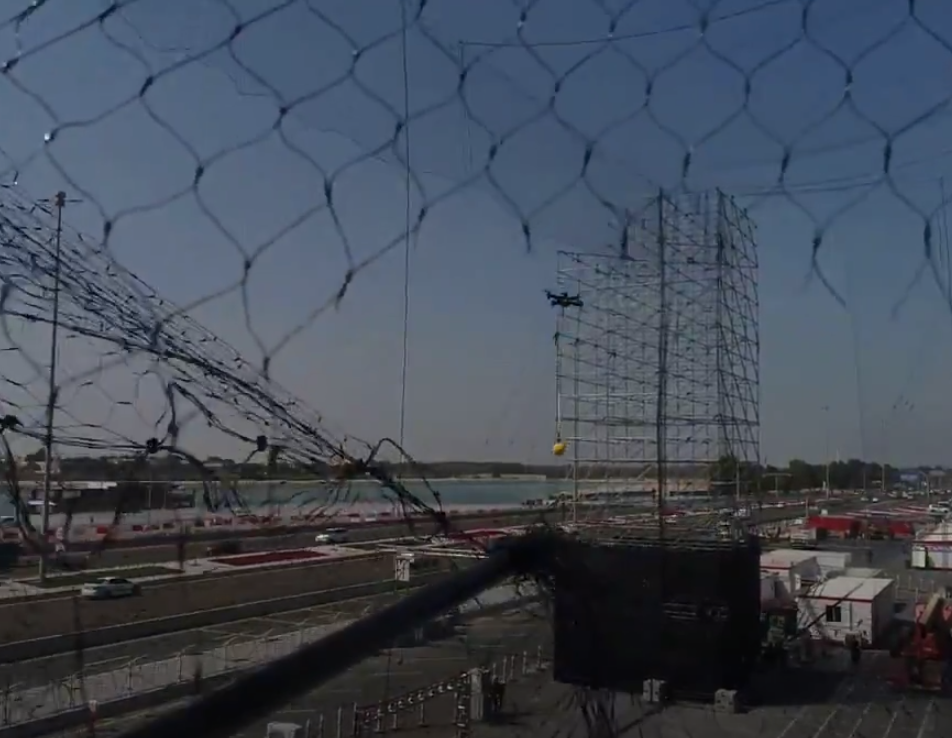}}\\
\caption{\label{fig:target-moving-away}Target ball moving away from UAV in Task 2. Netting is part of a UAV-mounted ball catcher. (a) LOS guidance issues a velocity command upwards, towards the target to intercept it on the next pass; (b) the LOS vector is pointing mostly forward; (c) the LOS vector is pointing downward as the target leaves view.}
\end{figure*}

As can be seen in Figure \ref{fig:target-moving-away}, as the frames progress, the LOS vector changes from pointing up towards the target, where the path of the target actually is, to having a smaller component upwards and instead pointing mostly forwards in the middle frame. By the last frame, the LOS vector is actually pointing downwards, away from the closest point in the target path. In general, an object moving away from a camera, in a direction perpendicular to the image plane, will project closer to the center of the image. Since the LOS vectors had a diminishing upward and even downward component as the target moved away, the UAV was never able to achieve a successful intercept.

% \begin{figure}[h!]
%     \centering
%     \includegraphics[width=0.6\textwidth]{figs/ball-moving-away.png}
%     \caption{Image geometry of ball moving away from camera.}
%     \label{fig:ball-moving-away}
% \end{figure}

The global search pattern successfully found the target and guidance towards it was geometrically correct. However, since there was no infrastructure developed for determining if the target was moving away or towards the UAV, the system was not able to adapt in the scenario presented here. Further analysis of the camera stream suggests that the UAV was less than 50cm from the target.

% TODO include results from object detection

% TODO report percent frames with ball detection

 \chapter{Conclusion}
\label{chap:conclusion}

% These results suggest that such forecasting or trajectory following methods may not be accurate if using just monocular camera data ...

\section{Discussion}

In this thesis, a variety of guidance algorithms were designed and implemented in simulated and real environments. Metrics used to assess the performance of each method include first-pass hit rate and pursuit duration. LOS Guidance methods were derived from the True Proportional Navigation algorithm, which comes from missile guidance literature. Implementation on a quadrotor allows greater control and maneuverability due to additional degrees of freedom. While TPN was shown to have the highest success rate, the results presented here also suggest that there are benefits to utilizing the additional degrees of freedom when operating on a quadrotor. For example, the Hybrid TPN-Heading method trades off between roll angle and yaw-rate control to implement lateral acceleration commands. This adopts the behavior of the TPN method when the target is centered in frame, but also changes heading to re-center the target when necessary. Hybrid TPN-Heading outperforms TPN at low target speeds. This, paired with the qualitative results presented, suggest that this method may outperform TPN with additional tuning and under certain conditions (e.g., large target motions including straight line and figure-8). TPN, the simplest method presented here, is preferred when a target is performing evasive maneuvers with high accelerations (e.g., tight turns). We thus arrive at the first key conclusion of this work: when implemented on a quadrotor, a hybrid model of yaw control and lateral acceleration control (through roll angle commands) can outperform the original True Proportional Navigation algorithm under certain conditions.

All of these methods relied only on the information from a monocular RGB camera onboard a quadrotor UAV. In Chapter \ref{chap:moving_objects} this camera was rigidly fixed to the UAV frame, and in Chapter \ref{chap:mbzirc} it was gimbal-mounted and set to point forward relative to the UAV heading. This sensor information was sufficient for good first-pass hit rate performance for some algorithms but not for others. As described in Section \ref{sec:results}, LOS Guidance methods such as TPN, PN-Heading, and Hybrid TPN-Heading generally had much higher hit rates than the Trajectory Following methods such as LOS' Trajectory and Forecasting Trajectory. In the simulated system, the vision was modeled realistically, so that a small target at long range suffered from a noisy LOS estimation due to pixel discretization. The trajectory methods therefore relied on smoothing of the LOS and LOS' to avoid generating trajectories with high variation. This smoothing, however, induced lag in the system and would sometimes lead to instability at high speeds or close ranges. For these reasons, we can arrive at the second key conclusion of this work: trajectory following methods generally require less noisy measurements of the target; with noisy measurements, simpler methods perform better.

Chapter \ref{chap:mbzirc} discussed the implementation of simple LOS following guidance on real systems, for both semi-stationary and moving targets, within a larger mission architecture. The global plan was successful in both parts of the mission (Tasks 1 and 2) in positioning the UAV such that it can effectively use LOS guidance to move towards the targets. In the semi-stationary targets case, even if the guidance did not result in a eliminated target on the first attempt, simply returning to the original point in space where the target detection was made and executing the guidance again always resulted in successful target elimination. In the moving target case, while the UAV was positioned correctly for LOS guidance, the inability to determine the target's flight direction caused inaccurate guidance and resulted in missing the target by an estimated distance of less than 50cm. From the competition results, we can arrive at a third key conclusion: in a scenario with repetitive target motions (semi-stationary or moving) simple LOS-direction guidance can effectively achieve target impact when within a larger system architecture including a global search plan and edge case infrastructure.

\section{Future Work}

This work opens questions regarding what guidance algorithms can be developed for quadrotors through only the use of monocular RGB camera imagery. As previously mentioned, it seems likely that further development of a hybrid method involving both proportional navigation and heading control might result in better performance. This may lead to higher hit rate, but also lower pursuit duration as the need to ``chase" the target might disappear as the UAV maintains the target in view in the initial approach. In addition, certain failure cases of the algorithms presented here suggest that more accurate controllers, particularly the thrust controller, might improve performance.

Due to the inability of Forecasting Trajectory to predict the 3D target position under moderate to high UAV speeds, suggesting that the UAV motion renders these estimates inaccurate, future work might limit the forecasting step to the 2D image plane. This would avoid the multiplicative nature of projecting inaccurate pixel estimates into 3D for determining target motion. Estimates of the target position in the image frame, paired with a LOS-based guidance method, might yield better results and greater stability at higher speeds.

\backmatter

%\renewcommand{\baselinestretch}{1.0}\normalsize

% By default \bibsection is \chapter*, but we really want this to show
% up in the table of contents and pdf bookmarks.
% \newcommand{\bibsection}{\chapter{\bibname}}
% \newcommand{\bibpreamble}{This text goes between the ``Bibliography'' header and the actual list of references}
% \bibsection
% \bibpreamble
\bibliography{bib} %your bib file

\end{document}